\documentclass{article} 
\usepackage[preprint]{colm2026_conference}

\usepackage{microtype}
\usepackage{hyperref}
\usepackage{url}
\usepackage{booktabs}

\usepackage{graphicx}
\usepackage{balance}  
\usepackage{amsmath}
\usepackage{algpseudocode}
\usepackage{latexsym}
\usepackage{multirow}
\usepackage{multicol}
\usepackage{color}
\usepackage{dashrule}
\usepackage{extarrows}
\usepackage{dsfont}
\usepackage{centernot}
\usepackage{hyperref}
\usepackage{natbib}
\usepackage{fancybox}
\usepackage[utf8]{inputenc} 
\usepackage[T1]{fontenc}    
\usepackage{hyperref}       
\usepackage{url}            
\usepackage{booktabs}       
\usepackage{amsfonts}       
\usepackage{amsmath}
\usepackage{nicefrac}       
\usepackage{microtype}      
\usepackage{xcolor} 
\usepackage{amsmath}
\usepackage{lipsum}
\usepackage{colortbl}
\usepackage{subcaption}
\usepackage{tabularx}
\usepackage{makecell}
\usepackage{wrapfig}
\usepackage{bm}
\usepackage{multirow}
\usepackage{xcolor} 
\usepackage{bbm}
\usepackage{mdframed}
\usepackage{enumitem}
\usepackage{xspace}

\usepackage{lineno}

\definecolor{darkblue}{rgb}{0, 0, 0.5}
\hypersetup{colorlinks=true, citecolor=darkblue, linkcolor=darkblue, urlcolor=darkblue}

\usepackage{listings}
\usepackage{xcolor}
\usepackage[ruled,vlined]{algorithm2e}
\usepackage{caption}
\usepackage{float} 
\usepackage{arydshln}
\usepackage{wrapfig} 
\usepackage{tcolorbox}
\tcbuselibrary{theorems}
\tcbuselibrary{most}
\usepackage[dvipsnames]{xcolor}
\usepackage{hyperref}       
\usepackage{url}            
\usepackage{booktabs}       
\usepackage{amsfonts}       
\usepackage{nicefrac}       
\usepackage{microtype}      
\usepackage{xcolor}         
\usepackage{amsmath}
\usepackage{multirow}
\usepackage{multicol}
\usepackage{colortbl}
\usepackage{tcolorbox}
\usepackage{wrapfig}
\usepackage{cleveref}
\usepackage{xspace}
\usepackage{graphicx}
\usepackage{url}
\usepackage{booktabs}
\usepackage{hyperref}
\usepackage{graphicx}
\usepackage{microtype}
\usepackage{multicol}
\usepackage{xcolor}   
\usepackage{multirow}
\usepackage{lipsum}
\usepackage{enumitem}
\usepackage{pifont}
\usepackage{colortbl}
\usepackage{tabularx}
\usepackage{caption} 
\usepackage{float}
\usepackage{amsmath,amsfonts,amssymb,bbm}
\usepackage{adjustbox}

\usepackage{lineno}

\definecolor{darkblue}{rgb}{0, 0, 0.5}
\hypersetup{colorlinks=true, citecolor=darkblue, linkcolor=darkblue, urlcolor=darkblue}
\definecolor{light-gray}{gray}{0.6}
\definecolor{front-color}{HTML}{F5FFFA}
\definecolor{Gray}{gray}{0.93}
\definecolor{customTeal}{RGB}{0, 128, 128} 
\definecolor{emphasisColor}{RGB}{255, 0, 0} 
\definecolor{oursBlue}{RGB}{51,202,246}

\definecolor{blue1}{HTML}{508AB2}
\definecolor{green2}{HTML}{BFF6BA}
\definecolor{closed_models}{RGB}{255, 219, 187}
\definecolor{open_models_below_4B}{RGB}{185, 235, 255}
\definecolor{open_models_7B_12B}{RGB}{39, 121, 1}

\definecolor{columngreen}{HTML}{D4F2D7} 
\definecolor{columnblue}{HTML}{C2E6F5} 
\definecolor{columnred}{HTML}{F5C2CC} 
\definecolor{columnpurple}{HTML}{E6D4F2} 
\definecolor{columnyellow}{HTML}{FBE0BC} 
\definecolor{columncyan}{HTML}{C2F5E6}  
\definecolor{columnpeach}{HTML}{F5D4C2} 
\newcommand{\model}{\textsc{G$^2$RPO}\xspace} 
\newcommand{\openvl}{{OpenVLThinkerV2}\xspace} %

\title{\openvl: A Generalist Multimodal Reasoning Model for Multi-domain Visual Tasks}

\author{Wenbo Hu \quad Xin Chen\quad Yan Gao-Tian \quad Yihe Deng \quad Nanyun Peng \quad  \\ \textbf{Kai-Wei Chang}
\\
University of California, Los Angeles (UCLA) \\
\{whu, kwchang\}@cs.ucla.edu
\\ 
\href{https://gordonhu608.github.io/openvlthinkerv2.github.io/}{\adjustbox{valign=c}{\includegraphics[height=1em]{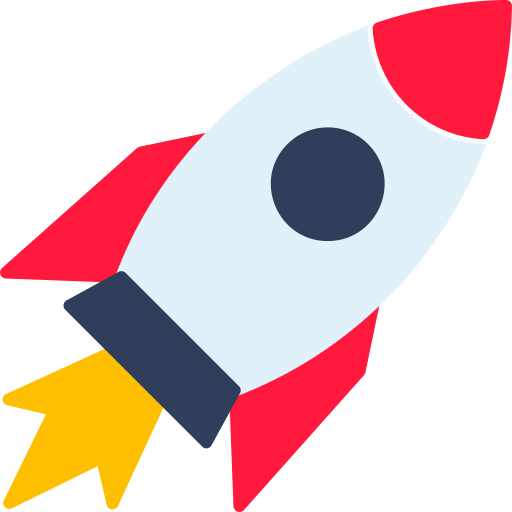}} Project Page} \quad
\href{https://github.com/uclanlp/openvlthinker}{\adjustbox{valign=c}{\includegraphics[height=1em]{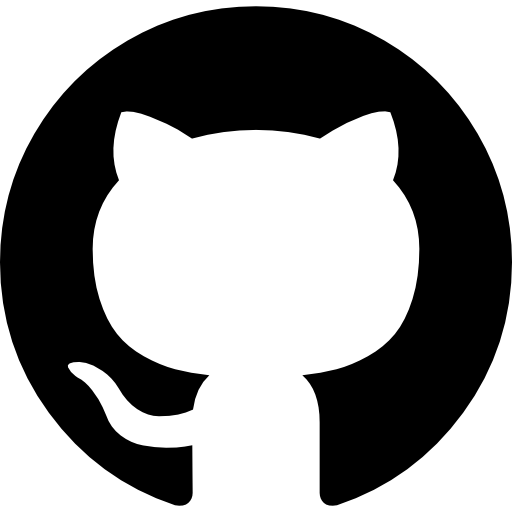}} GitHub} 
}

\begin{document}

\ifcolmsubmission
\linenumbers
\fi

\maketitle  

\begin{abstract}
Group Relative Policy Optimization (GRPO) has emerged as the de facto Reinforcement Learning (RL) objective driving recent advancements in Multimodal Large Language Models. However, extending this success to open-source multimodal generalist models remains heavily constrained by two primary challenges: the extreme variance in reward topologies across diverse visual tasks, and the inherent difficulty of balancing fine-grained perception with multi-step reasoning capabilities. To address these issues, we introduce \textbf{Gaussian GRPO (\model)}, a novel RL training objective that replaces standard linear scaling with non-linear distributional matching. By mathematically forcing the advantage distribution of any given task to strictly converge to a standard normal distribution, $\mathcal{N}(0,1)$, \model theoretically ensures inter-task gradient equity, mitigates vulnerabilities to heavy-tail outliers, and offers symmetric update for positive and negative rewards.
Leveraging the enhanced training stability provided by \model, we introduce two task-level shaping mechanisms to seamlessly balance perception and reasoning. First, response length shaping dynamically elicits extended reasoning chains for complex queries while enforce direct outputs to bolster visual grounding. 
Second, entropy shaping tightly bounds the model's exploration zone, effectively preventing both entropy collapse and entropy explosion.
Integrating these methodologies, we present \textbf{\openvl}, a highly robust, general-purpose multimodal model. Extensive evaluations across 18 diverse benchmarks demonstrate its superior performance over strong open-source and leading proprietary frontier models. 


\end{abstract}

\begin{figure}[h]
\vspace{-5mm}
  \centering
\includegraphics[trim=0cm 0cm 0cm 0cm, clip, width=\linewidth]{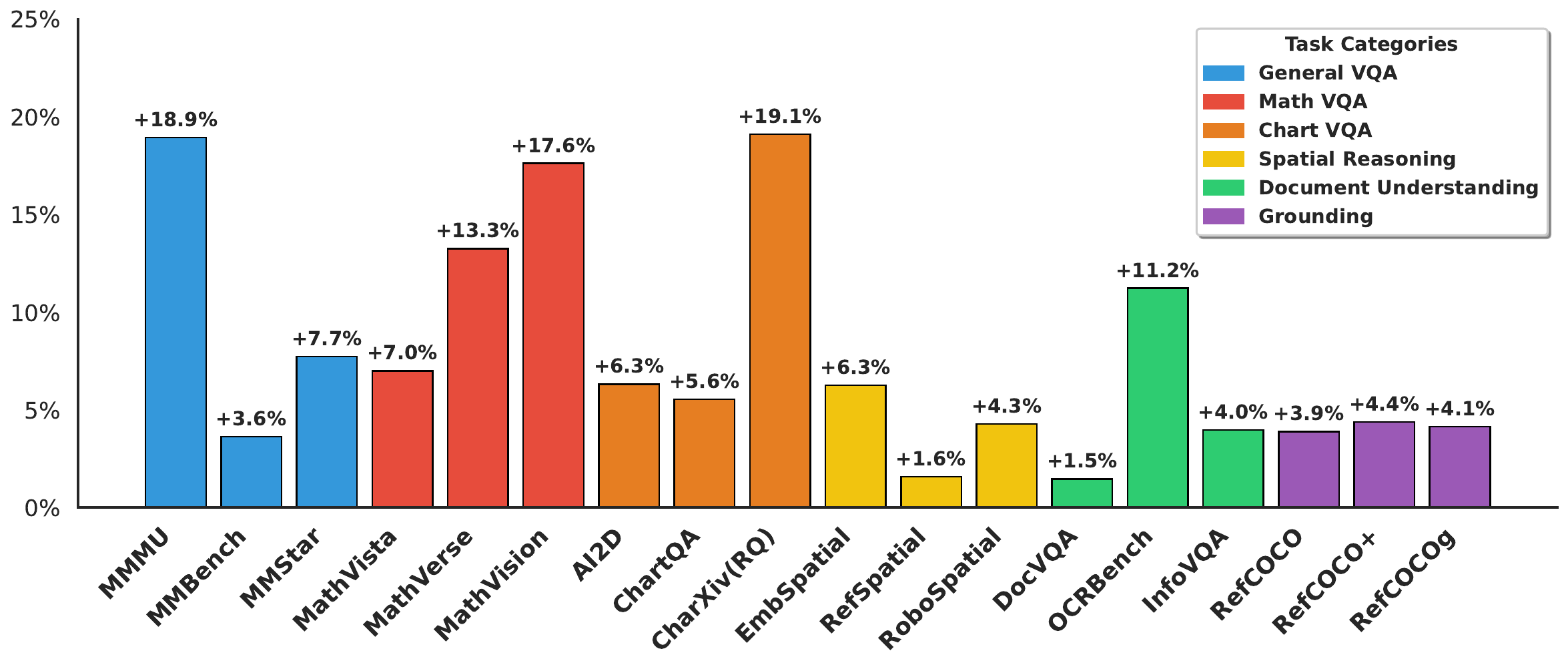}
  \caption{Performance improvement (relative) of \openvl over its baseline Qwen3-VL-Instruct-8B across diverse visual tasks.}
 \label{fig:teaser}
 \vspace{-5mm}
\end{figure}

\section{Introduction}
\vspace{-1mm}

Reinforcement Learning (RL) has emerged as a primary driver of recent advancements in Multimodal Large Language Models (MLLMs), significantly enhancing performance across domains ranging from complex visual reasoning to fine-grained object detection~\citep{qwen3vl, comanici2025gemini, singh2025openaigpt5card,liu2025visual,feng2025onethinkerallinonereasoningmodel, kimiteam2026kimik25visualagentic, seed2026seed18modelcardgeneralized}, encompassing the diverse spectrum of tasks illustrated in Figure~\ref{fig:teaser}.
However, the vast diversity of visual tasks imposes a significant challenge when optimizing them jointly during the MLLM post-training stage. The extreme variance in reward topologies---ranging from sparse, binary signals in math visual question answering (VQA) to dense, continuous Intersection-over-Union (IoU) scores in grounding tasks---creates significant intra- and inter-task update imbalances. This instability is particularly detrimental to the Group Relative Policy Optimization (GRPO)~\citep{guo2025deepseek} algorithm, rendering it highly susceptible to gradient explosion during large-scale training.

Standard GRPO suffers from intra-task imbalance because its sample-wise standard deviation normalization disproportionately favors low-variance rollouts~\citep{liu2025understanding, bereket2025uncalibrated, chu2025gpg,huang2025mapo}. Dr.GRPO~\citep{liu2025understanding} removes this normalization but inevitably causes inter-task imbalance, where high variance tasks dominate gradient update and low variance ones are suppressed. While recent methods like EMA-GRPO~\citep{feng2025onethinkerallinonereasoningmodel} mitigate this using task-wise moving averages of reward variance, they fundamentally rely on linear transformations. Since linear scaling merely matches the first two statistical moments (mean and variance) while preserving higher-order distributional shapes, it fails to guarantee true inter-task gradient equity and leaves the optimization vulnerable to structural pathologies like heavy-tail outliers.

\begin{figure}[t]
 \vspace{-10mm}
  \centering
\includegraphics[trim=0cm 53.2cm 28.5cm 0cm, clip, width=\linewidth]{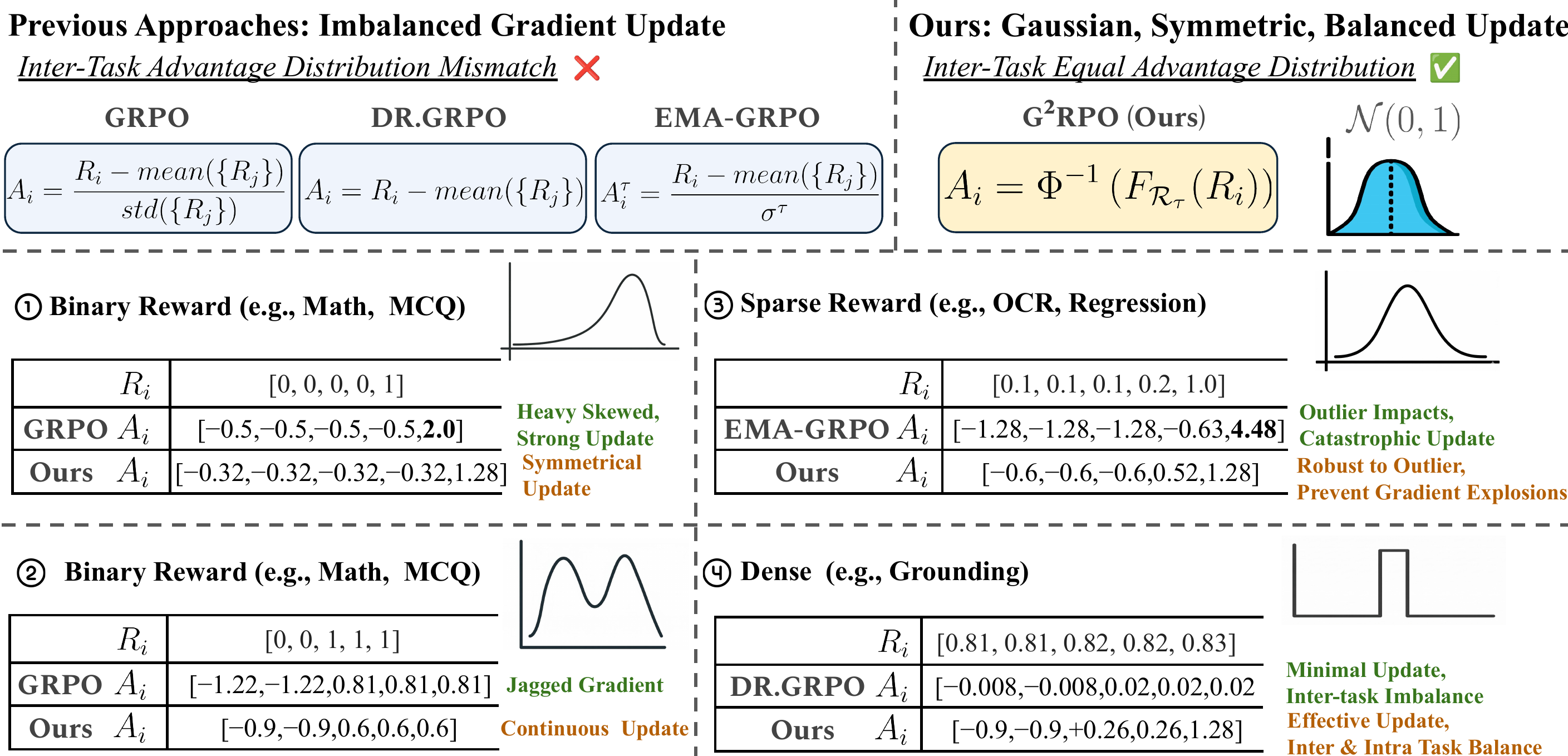}
  \caption{Comparison of advantage formulations against previous methods. By enforcing a Gaussian topology, \model provides 1) intrinsic robustness to outliers, 2) symmetric updates for positive and negative rewards, and 3) uniform variance across diverse tasks.}
 \label{fig:method}
 \vspace{-5mm}
\end{figure}

To overcome the statistical fragility of linear normalization, we propose \textbf{Gaussian GRPO (\model)}, which replaces scalar standardization with non-linear distributional matching. By utilizing 1D Optimal Transport---which admits a highly efficient closed-form solution via Cumulative Distribution Functions (CDFs)---\model strictly maps the empirical reward distribution of any given task directly to the standard normal distribution, $\mathcal{N}(0,1)$. As illustrated in Figure~\ref{fig:method}, enforcing this strict Gaussian topology mathematically caps outliers, smooths bimodal step-functions into symmetrically balanced tails, and theoretically ensures inter-task gradient equity.

Another major challenge in MLLM RL post-training is preserving robust capabilities in both fine-grained perception and multi-step reasoning~\citep{liu2025seeingbelievingprobingdisconnect,tian2025more,tu2025perceptionconsistencymultimodallargelanguage,yang2025learninglookdisentangledcurriculum}. While recent works~\citep{wang2025perceptionawarepolicyoptimizationmultimodal, tian2025more, zhang2026lessrightbidirectionalperceptual} attempt to enhance perception by perturbing visual inputs for auxiliary training objectives or by inserting explicit vision anchors, these approaches suffer from significant drawbacks. They require costly data annotations or incur substantial computational overhead from additional modules, severely limiting their scalability and leaving their efficacy unverified across diverse, multi-domain multimodal benchmarks. 

To address this with a simple and scalable approach, we frame this challenge as a multi-task balancing optimization between vision-centric tasks (e.g., OCR, visual grounding) and reasoning-centric tasks (e.g., math and science VQA). Alongside \model, we systematically analyze task-level performance, focusing specifically on the dynamics of response length and entropy loss during training. Observing a stark divergence in the optimization trajectories of  distinct task types, we introduce task-level \textit{response length and entropy shaping} to encourage stable and accelerated convergence. For response length, we explicitly elicit extended reasoning chains for complex queries while enforcing concise, direct outputs for vision-centric tasks, which effectively solves complex questions, mitigates hallucinations and bolsters visual grounding. Concurrently, our entropy shaping mechanism confines the model to an optimal exploration zone, preventing both entropy collapse (premature over-reliance on high-probability tokens) and entropy explosion (generation of incoherent text).

Integrating these methodologies, we introduce \textbf{\openvl}. We evaluate our model across 18 benchmarks spanning six major task categories: general science knowledge, mathematics, chart and document understanding, spatial reasoning, and visual grounding. Extensive experiments demonstrate that \openvl consistently achieves robust performance, establishing new state-of-the-art (SOTA) results among open-source models. For instance, \openvl achieves $71.6\%$ on MMMU and $79.5\%$ on MathVista, surpassing GPT-4o by a significant margin. Furthermore, across six distinct benchmarks evaluating document understanding and spatial reasoning, \openvl significantly outperforms proprietary frontier models, including GPT-5 and Gemini 2.5 Pro.

Our main contributions are summarized as follows:
\vspace{-2mm}
\begin{itemize}[align=right,itemindent=0em,labelsep=2pt,labelwidth=1em,leftmargin=*,itemsep=0em]
\item We propose \textbf{\model}, a novel RL training objective that replaces linear scaling with non-linear distributional matching. By mathematically forcing each task's advantage distribution to converge to $\mathcal{N}(0,1)$, \model theoretically ensures inter-task gradient equity and systematically mitigates vulnerabilities to structural pathologies like heavy-tail outliers.

\item We introduce \textbf{task-level response length and entropy shaping} mechanisms to balance perception and multi-step reasoning. These dynamic bounds encourages early response length convergence and effectively preventing both entropy collapse and explosion.

\item We present \textbf{\openvl}, a robust, general-purpose multimodal model. Extensive evaluations across 18 benchmarks demonstrate its superior performance, establishing new SOTA results and consistently outperforming leading proprietary frontier models.
\end{itemize}

\section{Method}
\vspace{-1mm}
\label{sec:method}

\subsection{Preliminary}
\label{subsec:grpo}
\vspace{-1mm}
\textbf{Group Relative Policy Optimization (GRPO).}
We model an autoregressive language model as a stochastic policy $\pi_\theta$. For a given query $q \sim \mathcal{D}$, GRPO samples a group of $G$ responses $\mathcal{G} = \{y_1,\ldots,y_G\}$ from the behavior policy $\pi_{\theta_\text{old}}$ and computes their scalar rewards $\{R_1,\ldots,R_G\}$. It then maximizes the following token-level objective:
\begin{align}
\label{equ:grpo}
\mathcal{J}_\text{GRPO}(\theta) = \mathbb{E}_{ q \sim \mathcal{D},\, \{y_i\}_{i=1}^G \sim \pi_{\theta_\text{old}} }
\left[ \frac{1}{G} \sum_{i=1}^{G} \frac{1}{|y_i|} \sum_{t=1}^{|y_i|} 
\min \left( r_{i,t}(\theta) \widehat{A}_{i}, \, \text{clip} \left( r_{i,t}(\theta), 1 - \varepsilon, 1 + \varepsilon\right) \widehat{A}_{i} \right)
\right]
\end{align}
where $r_{i,t}(\theta)$ is the probability ratio between the new and old policy, and $\varepsilon$ is the clipping range. The advantage $\widehat{A}_{i}$ is defined by standardizing the rewards within the prompt group:
\begin{equation}
    \widehat{A}_{i}^{\text{GRPO}} = \frac{ R_i - \mu_{\mathcal{G}} }{ \sigma_{\mathcal{G}} + \epsilon }
\label{eq:grpo_adv}
\end{equation}
where $\mu_{\mathcal{G}}$ and $\sigma_{\mathcal{G}}$ are the group's empirical mean and standard deviation.

\textbf{Multi-Task Normalization.}
\label{subsec:ema_grpo}
While GRPO is highly effective for single-task alignment, its localized standard deviation $\sigma_{\mathcal{G}}$ destabilizes optimization in multi-task scenarios where reward scales differ drastically. EMA-GRPO addresses this inter-task imbalance by scaling the advantage using a historically tracked, task-specific standard deviation $\sigma_{\tau}^{(t)}$:
\begin{equation}
    \sigma_{\tau}^{(t)} = \alpha \sigma_{\tau}^{(t-1)} + (1-\alpha) \sigma_{\mathcal{G}}, \quad \quad \widehat{A}_i^{\text{EMA-GRPO}} = \frac{R_i - \mu_{\mathcal{G}}}{\sigma_{\tau}^{(t)} + \epsilon}
\end{equation}

However, these approaches do not fundamentally resolve the issue of imbalanced gradient updates across diverse tasks. Because standard normalization relies on a \textit{linear transformation}, it strictly preserves the original shape of the reward distribution. This limitation leaves the optimization process vulnerable to severe topological pathologies. For example, as illustrated in Figure~\ref{fig:method}'s scenarios 1 and 3, it is susceptible to \textit{heavy-tail outliers}, where a single anomalously high reward artificially inflates the exponential moving average (EMA) variance over time, thereby suppressing the learning signals for subsequent normal responses. In addition, EMA-GRPO requires extra hyperparameter tuning and introduces a \textit{momentum lag} where the fixed decay rate $\alpha$ struggles to adapt to sudden policy breakthroughs, inadvertently feeding the model stale advantage signals.

\begin{figure}[t]
\vspace{-10mm}
  \centering
\includegraphics[width=\linewidth]{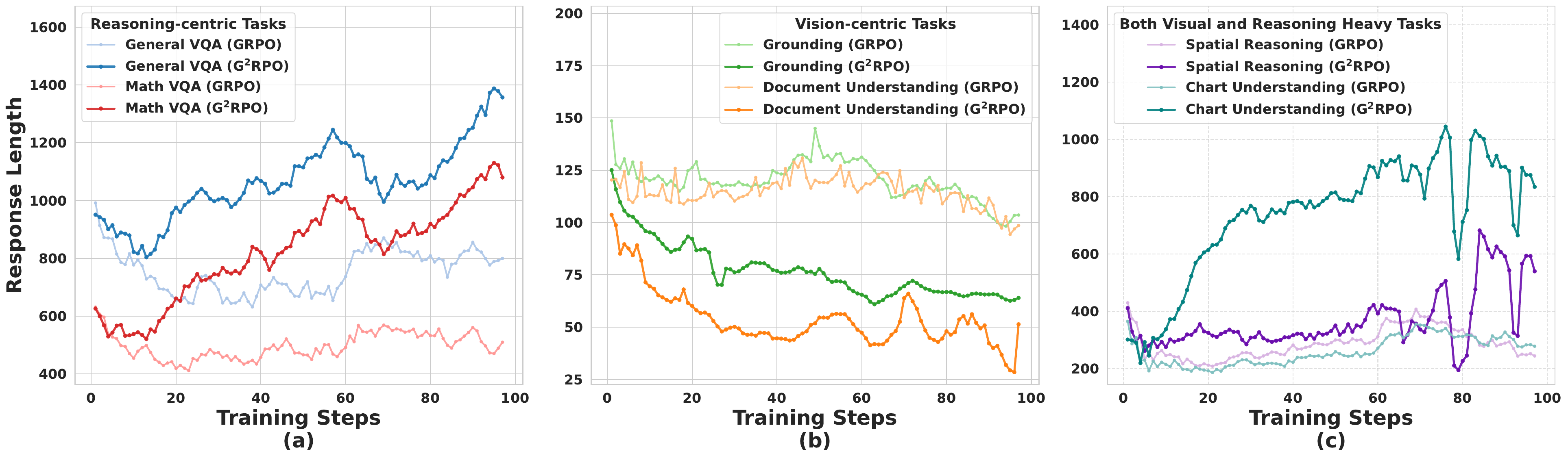}
  \caption{Comparison of response length dynamics during training. \model effectively encourage early convergence. a) It scales up reasoning length for complex question. b) It reduces overthinking for visual-centric tasks, enhancing perceptual grounding and mitigating hallucinations. c) For both reasoning and perception heavy tasks, the generation length stabilizes within an optimal range, effectively balancing both capabilities.}
 \label{fig:study_length}
 \vspace{-5mm}
\end{figure}

\subsection{Gaussian GRPO (\model)}
\label{subsec:g_grpo}
\vspace{-1mm}

To overcome the statistical fragility of linear moment-matching, we abandon scalar standardization in favor of non-linear distributional matching. Intuitively, an optimal advantage distribution should be {intrinsically robust to outliers}, {symmetric across positive and negative rewards}, and maintain {uniform variance across highly diverse tasks}. These structural requirements naturally motivate the adoption of a Gaussian distribution. 
Therefore, we propose \textbf{Gaussian GRPO (\model)}, which formulates advantage estimation as an Optimal Transport problem. Specifically, our objective is to find a transport map that strictly maps the empirical reward distribution of any task directly to a well-behaved target distribution, namely the standard normal distribution $P_{\mathcal{N}} \equiv \mathcal{N}(0,1)$.

Given a set of empirical rewards $\mathcal{R}_{\tau} = \{R_1, \dots, R_N\}$ for a specific task $\tau$, let $P_{\mathcal{R}_{\tau}}$ denote its empirical distribution, our goal is to find a mapping function $\Psi$ that transports this empirical distribution to $\mathcal{N}(0,1)$. This can be achieved by minimizing the Wasserstein-2 ($W_2$) distance between $P_{\mathcal{R}_{\tau}}$ and $P_{\mathcal{N}}$, defined under the squared Euclidean cost function:
\begin{equation}
    W_2^2(P_{\mathcal{R}_{\tau}}, P_{\mathcal{N}}) = \inf_{\gamma \in \Pi(P_{\mathcal{R}_{\tau}}, P_{\mathcal{N}})} \int_{\mathbb{R} \times \mathbb{R}} |x - y|^2 \, d\gamma(x,y)
\end{equation}
where $\Pi(P_{\mathcal{R}_{\tau}}, P_{\mathcal{N}})$ denotes the collection of all joint distributions with marginals $P_{\mathcal{R}_{\tau}}$, $P_{\mathcal{N}}$. 


In 1-dimensional space, this unique optimal transport map admits a highly efficient closed-form solution via the Cumulative Distribution Functions (CDFs). By evaluating this map, \model mathematically neutralizes extreme skewness, bimodal splits, and outliers. Instead of standardizing via mean and variance, \model assigns advantages by mapping the relative rank of the response directly to the inverse CDF of the target normal distribution:
\begin{equation}
    \widehat{A}_i^{\text{Ours}} = \Psi(R_i, \mathcal{R}_{\tau}) = \Phi^{-1}\left( F_{\mathcal{R}_{\tau}}(R_i) \right)
\end{equation}
where $\Phi^{-1}$ is the quantile function (inverse CDF) of $\mathcal{N}(0,1)$, and $F_{\mathcal{R}_{\tau}}$ is the empirical CDF of the task's rewards. We provide a step-by-step pytorch-style pseudocode implementation of \model, alongside its underlying mathematical formulations, in Algorithm \ref{alg:ggrpo}.


\textbf{Discussion.} 
By enforcing a strict Gaussian topology, \model isolates the policy from anomalous reward spikes by mathematically capping outliers at the highest quantile. As illustrated in Figure~\ref{fig:method}, for binary rewards, it gracefully converts bimodal step-functions into smooth, symmetrically balanced Gaussian tails. Because every task's advantage distribution is mathematically forced to converge to $\mathcal{N}(0,1)$, \model theoretically ensures inter-task equity without requiring rigid momentum hyperparameters.


\begin{algorithm}[t]
\caption{\model: Distributional Matching via 1D Optimal Transport}
\label{alg:ggrpo}
\textbf{Objective:} Map empirical task rewards $\mathcal{R}_{\tau} = \{R_1, \dots, R_N\}$ to $\mathcal{N}(0,1)$. \\
\vspace{1mm}

\noindent\textbf{Step 1: Rank Rewards.} Compute the uniform probability $p_i$ based on the relative rank of each reward.
\begin{equation}
    p_i = \frac{\text{rank}(R_i) - 0.5}{N}
\end{equation}
\vspace{-3mm}
\begin{lstlisting}[language=Python, numbers=none, breaklines=true, basicstyle=\ttfamily\small, keywordstyle=\color{blue}]
# Sort empirical rewards and track original indices
sorted_rewards, indices = torch.sort(rewards)
ranks = torch.arange(1, N + 1, device=device, dtype=torch.float32)
probabilities = (ranks - 0.5) / N
\end{lstlisting}

\vspace{1mm}
\noindent\textbf{Step 2: Quantile Mapping.} Map $p_i$ to the inverse CDF (quantile function) of $\mathcal{N}(0,1)$, denoted as $\Phi^{-1}$.
\begin{equation}
    \Psi(R_i, \mathcal{R}_{\tau}) = \Phi^{-1}\left( p_i \right) = \sqrt{2} \operatorname{erfinv}(2p_i - 1)
\end{equation}

\begin{lstlisting}[language=Python, numbers=none, breaklines=true, basicstyle=\ttfamily\small, keywordstyle=\color{blue}]
# Generate target quantiles from Standard Normal N(0,1)
target_quantiles = math.sqrt(2.0) * torch.erfinv(2.0 * probabilities - 1.0)
target_quantiles = target_quantiles.to(dtype)
\end{lstlisting}

\vspace{1mm}
\noindent\textbf{Step 3: Tie-Breaking Strategy.} To ensure identical behaviors receive identical learning signals, assign the mean of the target quantiles to all identical values, where $\mathcal{K}_{R_i}$ is the set of indices $j$ such that $R_j = R_i$.
\begin{equation}
    \widehat{A}_i^{\text{Ours}} = \Psi_{\text{tied}}(R_i, \mathcal{R}_{\tau}) = \frac{1}{|\mathcal{K}_{R_i}|} \sum_{j \in \mathcal{K}_{R_i}} \Psi(R_j, \mathcal{R}_{\tau})
\end{equation}
\vspace{-4mm}
\begin{lstlisting}[language=Python, numbers=none, breaklines=true, basicstyle=\ttfamily\small, keywordstyle=\color{blue}]
# Handle Ties: Average the quantiles for identical rewards
unique_rewards, inverse_indices = torch.unique(sorted_rewards, return_inverse=True)
if unique_rewards.shape[0] < N:
    tied_quantiles = torch.zeros_like(unique_rewards, dtype=dtype)
    tied_quantiles.scatter_add_(0, inverse_indices, target_quantiles)
    counts = torch.bincount(inverse_indices).to(dtype)
    tied_quantiles = tied_quantiles / counts
    target_quantiles = tied_quantiles[inverse_indices]
# Scatter advantages back to original tensor ordering
advantages = torch.zeros_like(rewards)
advantages[indices] = target_quantiles
\end{lstlisting}
\end{algorithm}

\subsection{Task-level Length and Entropy Shaping}
\vspace{-1mm}

A primary challenge in advancing RL post-training for MLLMs lies in preserving robust capabilities across both fine-grained perception and multi-step reasoning~\citep{liu2025seeingbelievingprobingdisconnect,tian2025more,tu2025perceptionconsistencymultimodallargelanguage,yang2025learninglookdisentangledcurriculum}. We frame this as a multi-task balancing optimization, distinguishing between vision-centric tasks (e.g., OCR, visual grounding), reasoning-centric tasks (e.g., math and science VQA), and hybrid tasks requiring both modalities (e.g., chart reasoning). To this end, we first analyze the task-level performance dynamics of these distinct categories, with a specific focus on the evolution of response length and entropy loss during training.

\textbf{Task-Level Response Length Shaping.} As illustrated in Figure~\ref{fig:study_length}, we observe distinct trajectories in response length dynamics during GRPO training. For reasoning-centric tasks, the model initially decreases response length as it adapts to the training data distribution, before ultimately converging to longer, more elaborate reasoning chains. Conversely, for vision-centric tasks, the model consistently reduces response length, suggesting an optimization toward concise outputs that directly enhance perceptual grounding without generating unnecessary, potentially hallucinated reasoning steps.

Motivated by these divergent trends, we explicitly shape the expected response length on a per-task basis to accelerate convergence and jointly enhance both perception and reasoning capabilities. We propose a task-level length shaping mechanism that applies a customized trapezoidal reward envelope, rewarding the model when its generation length falls within an optimal target range while softly penalizing excessively short or long responses. 

For a given task, let $|y|$ denote the token length of the model's response $y$. We define four task-specific threshold hyperparameters: the absolute minimum valid length $L_{\text{min}}$, the start of the optimal length plateau $L_{\text{low}}$, the end of the optimal plateau $L_{\text{high}}$, and the absolute maximum valid length $L_{\text{max}}$. The length reward $R_{\text{length}}(y)$ is mathematically formulated as:

\begin{equation}
R_{\text{length}}(y) =
\begin{cases}
0, & |y| < L_{\text{min}} \text{ or } |y| > L_{\text{max}} \\
\frac{|y| - L_{\text{min}}}{L_{\text{low}} - L_{\text{min}}}, & L_{\text{min}} \le |y| < L_{\text{low}} \\
1, & L_{\text{low}} \le |y| \le L_{\text{high}} \\
\frac{L_{\text{max}} - |y|}{L_{\text{max}} - L_{\text{high}}}, & L_{\text{high}} < |y| \le L_{\text{max}}
\end{cases}
\label{eq:length_reward_envelope}
\end{equation}


\begin{figure}[t]
\vspace{-10mm}
  \centering
\includegraphics[width=\linewidth]{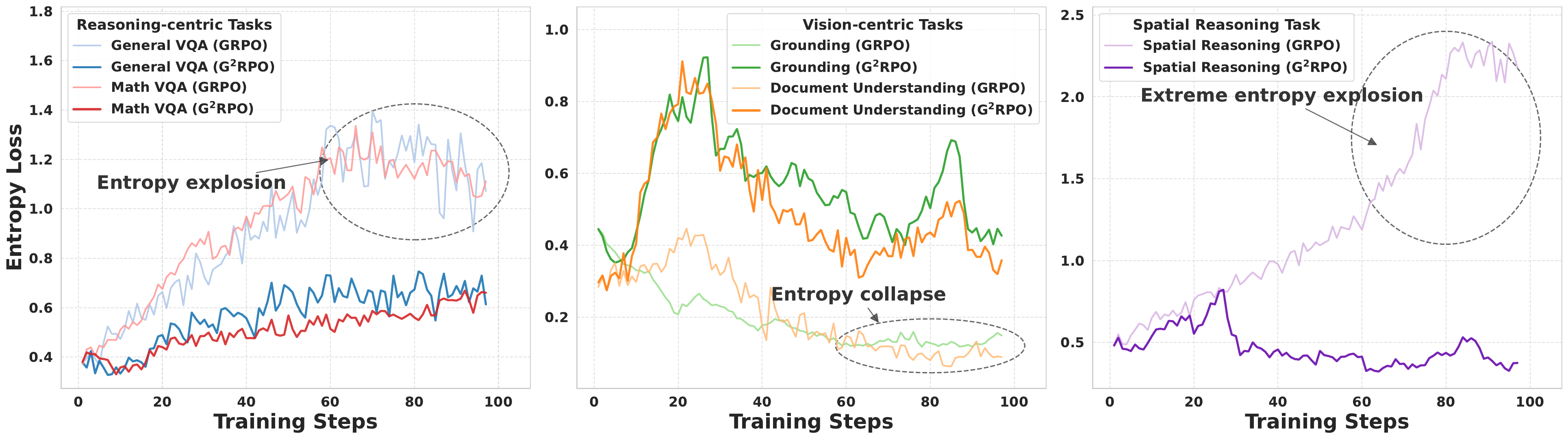}
  \caption{Effect of task-level entropy shaping. \model effectively prevents entropy explosion for reasoning-centric tasks and OOD task (spatial reasoning) while concurrently mitigating entropy collapse in vision-centric tasks.}
 \label{fig:study_entropyh}
 \vspace{-5mm}
\end{figure}

\textbf{Task-Level Entropy Shaping.} In a multi-task reinforcement learning setting, varying task difficulties inevitably induce divergent exploration patterns. As illustrated in Figure~\ref{fig:study_entropyh}, we observe that for reasoning-centric tasks, the model tends to artificially inflate entropy, leading to \textit{entropy explosion} where it explores incoherent tokens. Conversely, vision-centric tasks are highly susceptible to \textit{entropy collapse}, wherein the model over-exploits high-probability tokens and prematurely abandons necessary exploration. Furthermore, the phenomenon of entropy explosion is drastically exacerbated in complex or out-of-distribution (OOD) tasks, indicating a detrimental inclination to sample low-probability regions of the action space.

To mitigate these training instabilities, we propose a task-level entropy shaping mechanism. Analogous to our response length shaping, we impose a strict regularization envelope to bound the model's exploration within an optimal, task-specific zone. Let $H_{\text{task}}$ denote the average entropy loss for a given task, computed over the generated negative log-probabilities. We define a target exploration interval bounded by a minimum entropy threshold, $H_{\text{min}}$, to prevent collapse, and a maximum entropy threshold, $H_{\text{max}}$, to prevent explosion. We formulate the entropy regularization loss, $\mathcal{L}_{\text{ent\_reg}}$, as a margin-based penalty that applies a linear correction only when the entropy falls outside this optimal envelope:
\begin{equation}
\mathcal{L}_{\text{ent\_reg}} = \max(0, H_{\text{task}} - H_{\text{max}}) + \max(0, H_{\text{min}} - H_{\text{task}})
\label{eq:entropy_reg}
\end{equation}

This regularization term is then added to the final optimization objective, scaled by a weighting hyperparameter $\lambda_{\text{ent}}$, ensuring the model maintains a stable balance between exploration and exploitation across highly diverse task topologies.

\textbf{Discussion.} While our response length and entropy shaping mechanisms introduce specific hyperparameters (e.g., $L_{\text{min}}$, $H_{\text{min}}$),
we argue that these boundaries can be intuitively selected to guide the model's behavioral trajectory---such as encouraging longer reasoning chains for complex tasks---without the need for delicate tuning. Although our current implementation relies on simple empirical observations to establish a general optimization trend, it already yields substantial performance improvements. We leave the systematic exploration and automated search of these hyperparameters to future work. We emphasize that even coarse, trend-based shaping is sufficient to achieve robust and significant gains.

\begin{table*}[t!]
    \vspace{-10mm}
    \centering
    \footnotesize
    \setlength{\tabcolsep}{0.5em}
    \renewcommand{\arraystretch}{1.15} 
    \resizebox{\textwidth}{!}{
    \begin{tabular}{l | ccc | ccc | ccc} 
    \hline
    \multicolumn{1}{c|}{} 
     & \multicolumn{3}{c|}{\cellcolor{columnblue}\textbf{General VQA}}
     & \multicolumn{3}{c|}{\cellcolor{columnred}\textbf{Math VQA}} 
     & \multicolumn{3}{c}{\cellcolor{columnyellow}\textbf{Chart VQA}} \\ 
     \cline{2-10}
     \textbf{Model} & \textbf{MMMU} & \textbf{MMBench} & \textbf{MMStar} & \textbf{MathVista} & \textbf{MathVerse} & \textbf{MathVision} & \textbf{AI2D} & \textbf{ChartQA} & \textbf{CharXiv(RQ)} \\ 
    \hline
    GPT-4o~\citep{hurst2024gpt} & 70.7 & 84.3 & 65.1 & 63.8 & 41.2 & 30.4& 84.9 & 86.7 & 47.1\\
    Gemini 2.5 Pro~\citep{comanici2025gemini} &  81.7 &  90.1 & 77.5 & 82.7 & 82.9 & 73.3 & 88.4 & 83.3 & 67.9\\
    \hline
    MM-Eureka-7B~\citep{meng2025mm} & 57.3 & 87.7 & 64.4 & 73.0 & 50.3  & 26.9 & 84.1 & 77.3 & 39.5 \\ 
    OpenVLThinker-7B~\citep{deng2025openvlthinker} & 55.8 & 87.8 & 59.1 & 70.2 & 47.9 & 25.3 & 81.8 & 85.7 & 39.3 \\ 
    VL-Rethinker-7B~\citep{wang2025vl} & 56.7 & 87.9 & 62.7 & 74.9 & 54.2 & 32.3 & 80.8 & 84.7 & 39.8  \\ 
   
    Vision-G1~\citep{zha2025visiong1generalvisionlanguage} & 53.4 & 88.0 & 63.1  &76.1 & 50.0 & 31.3  & 82.1 & 85.6 & 41.0  \\
    ARES-7B~\citep{chen2025ares} & 67.9& - & 65.3 & 74.6 & 56.5 & 51.9 & - & - & 47.0 \\ 

    OVR-7B~\citep{ovrwei2025openvisionreasonertransferring}  & - & 86.6 & 62.7 & 72.1 & 54.6 & 51.8 & - & - & 44.5  \\ 
    VisionZero~\citep{wang2026visionzeroscalablevlmselfimprovement} & 58.8 & - & 65.8 & 73.1 & 52.1 & 28.5 & 84.8 & 86.3 & -   \\ 
   
    OneThinker-8B~\citep{feng2025onethinkerallinonereasoningmodel} &  70.6 & 86.6 & 70.6 & 77.6 & 64.3 &48.3 & 85.2 & 76.4 & 44.0 \\ 
    \hline
    Qwen3-VL-Instruct-8B &60.2 & 85.1 & 68.5 & 74.2 & 58.1 &45.4 & 82.3 & 82.8 & 44.5 \\
    Qwen3-VL GRPO &  69.7 & 86.0 & 71.7 & 78.1 & 64.7& 52.6 & 85.1 & 82.4 & 50.5 \\
    Qwen3-VL GDPO & 69.8 & 86.1 & 71.5 & 78.0 & 64.9 & 49.7 & 85.1 & 82.4  &51.6  \\
  
    \hline
    \rowcolor{green!5}
    \textbf{\openvl (Ours)} & \textbf{71.6}  & \textbf{88.2} & \textbf{73.8} & \textbf{79.5} & \textbf{65.8}   & \textbf{53.4} & \textbf{87.5}  &  \textbf{87.4} & \textbf{53.0} \\
    \hline
    \end{tabular}
    } 
    \caption{Performance comparison of various models across visual reasoning benchmarks. We report our reproduced results for Qwen3-VL-Instruct-8B under the same setting. Our model consistently outperforms strong open-source baselines.}
    \label{tab:vqa-benchmarks_2}
    \vspace{-5mm}
\end{table*}

\section{Experiments}
\vspace{-1mm}
\subsection{Setup}
\vspace{-1mm}

\textbf{Training Details.} We train our model on AWS Trainium instances (specifically, Trn1.32xlarge). Our RL optimization is initialized from the Qwen3-VL-Instruct-8B~\citep{qwen3vl} and trained using a filtered subset of the OneThinker-600k dataset~\citep{feng2025onethinkerallinonereasoningmodel}. We optimize for a single epoch using AdamW, with a batch size of 128 and a learning rate of $2\times10^{-6}$. The maximum generation length is capped at 4096 tokens. Following the practices outlined in~\citet{yu2025dapo}, we disable KL regularization and apply dynamic data filtering, actively discarding rollouts that are uniformly correct or incorrect to maintain high-quality gradient signals. The entire training process requires approximately three days.

\textbf{Benchmarks.} To comprehensively evaluate our model, we adopt a variety of benchmarks across a wide-range of tasks. For {visual reasoning}, we evaluate on general multimodal VQA~\citep{yue2024mmmu, liu2024mmbench, chen2024we} and math-focused VQA~\citep{lu2023mathvista, zhang2024mathverse, wang2024measuring}. For {vision-centric perception}, we assess document understanding~\citep{mathew2021docvqa, mathew2022infographicvqa, OCRBench} and visual grounding~\citep{kazemzadeh2014referitgame, yu2016modeling}. For tasks that demand both rigorous visual perception and multi-step reasoning, we test on chart understanding~\citep{kembhavi2016diagram, masry-etal-2022-chartqa, wang2024charxiv} and spatial reasoning~\citep{du-etal-2024-embspatial, zhou2025roboreferspatial, song2025robospatial}. We report our reproduced results for Qwen3-VL-Instruct.
All models are evaluated following prior Qwen3-VL's default generation hyperparameters.

\subsection{Main Results}
\vspace{-1mm}

\textbf{Visual Reasoning Tasks.} As illustrated in Table~\ref{tab:vqa-benchmarks_2}, we evaluate \openvl across a diverse suite of tasks encompassing general scientific knowledge, mathematics, chart understanding, and complex multimodal reasoning. Compared to strong open-source baselines, including the recent generalist model OneThinker-8B, \openvl achieves superior performance across all evaluated benchmarks. Under controlled experimental conditions (utilizing identical training data and compute resources), our model consistently outperforms both standard GRPO and GDPO variants, underscoring the efficacy of \model. Notably, \openvl reaches $71.6\%$ on MMMU, $88.2\%$ on MMBench, and $73.8\%$ on MMStar, surpassing GPT-4o, as well as $87.4\%$ on ChartQA, which exceeds the performance of Gemini 2.5 Pro.

\begin{table*}[t!]
\vspace{-10mm}
    \centering
    \footnotesize
    \renewcommand{\arraystretch}{1.2} 

    \begin{subtable}{0.485\textwidth}
        \centering
        \resizebox{\linewidth}{!}{
        \begin{tabular}{l | ccc }
        \hline
        \multicolumn{1}{c|}{} & \multicolumn{3}{c}{\cellcolor{columncyan}\textbf{(a) Document Understanding}} \\
        \hline
        \textbf{Model}  & \textbf{DocVQA} & \textbf{OCRBench} & \textbf{InfoVQA} \\ 
        \hline
        GPT-5~\citep{singh2025openaigpt5card} & 91.5 &  810& 79.0 \\
        Gemini 2.5 Pro~\citep{comanici2025gemini} & 92.6 & 866 & 84.2 \\
        \hline
        DocThinker-7B~\citep{yu2025docthinkerexplainablemultimodallarge} & 78.8 & - & 52.3 \\
        VisionThink~\citep{yang2025visionthink} & 94.4 & 808 & - \\
        
        DeepEyesV2~\citep{hong2025deepeyesv2} & - & 882 & - \\ 
        RegionDoc-R1~\citep{wang2025regiondocr1} & 95.3 & - & 83.2 \\ 
        VisionZero~\citep{wang2026visionzeroscalablevlmselfimprovement} & 95.2 & 885 & 82.3 \\
        OneThinker-8B~\citep{feng2025onethinkerallinonereasoningmodel} &95.0 & 833 & 84.8  \\
        \hline
        Qwen3-VL-Instruct  & 95.3 & 819 & 83.1 \\
        Qwen3-VL GRPO & 95.9 & 875 & 84.9 \\
        Qwen3-VL GDPO & 95.6 & 897 & 83.5  \\
        \hline
        \rowcolor{green!5}
        \textbf{\openvl (Ours)} & \textbf{96.7}& \textbf{911} &  \textbf{86.4} \\
        \hline
        \end{tabular}
        }
    \end{subtable}\hfill
    \begin{subtable}{0.515\textwidth}
        \centering
        \resizebox{\linewidth}{!}{
        \begin{tabular}{l | ccc }
        \hline
        \multicolumn{1}{c|}{} & \multicolumn{3}{c}{\cellcolor{columnpeach}\textbf{(b) Spatial Reasoning}} \\
        \hline
        \textbf{Model} & \textbf{EmbSpatial} & \textbf{RefSpatial} & \textbf{RoboSpatial} \\ 
        \hline
        GPT-5~\citep{singh2025openaigpt5card} & 82.9 & 23.8 & 53.5 \\
        Gemini 2.5 Pro~\citep{comanici2025gemini} & 79.1 & 36.5 & 47.5 \\
        \hline
        SpatialRGPT-8B~\citep{cheng2024spatialrgpt} & 59.6 & - & 66.7 \\
        SpaceLLaVA-13B~\citep{foutter2025spacellavavisionlanguagemodeladapted} & 49.4 & 3.25 & 61.0 \\ 
        RoboBrain2.0-7B~\citep{baairobobrainteam2025robobrain20technicalreport} & - & 32.5 & 59.6  \\ 
        RoboRefer-8B-SFT~\citep{zhou2025roboreferspatial} & - & 48.4 & 58.3 \\
        G$^2$VLM~\citep{hu2025g2vlmgeometrygroundedvision} & 62.4 & 43.5 & 62.7 \\ 
        OneThinker-8B~\citep{feng2025onethinkerallinonereasoningmodel} &79.9  & 38.9 & 61.7\\ 
        \hline
        Qwen3-VL-Instruct  & 78.2 & 43.9 & 60.6 \\ 
        Qwen3-VL GRPO & 79.9 & 43.3 & 61.7 \\ 
        Qwen3-VL GDPO & 79.9 & 43.0 & 61.7 \\
        \hline 
        \rowcolor{green!5}

        \textbf{\openvl (Ours)} & \textbf{83.1} & \textbf{44.6} & \textbf{63.2} \\
        \hline
        \end{tabular}
        }
    \end{subtable}

    \caption{Results on (a) document understanding and (b) spatial reasoning benchmarks. We compare against both general multimodal models and task-specific expert baselines.}
    \label{tab:doc-spatial-benchmarks}
    \vspace{-5mm}
    
\end{table*}

\textbf{Document Understanding Tasks.} We evaluate the vision-centric capabilities of \openvl, particularly document understanding, as detailed in Table~\ref{tab:doc-spatial-benchmarks}(a). Our model achieves state-of-the-art performance, outperforming both strong open-source and leading proprietary models across all three benchmarks. Notably, \openvl attains a score of 911 on OCRBench, surpassing DeepEyesV2,  which is a specialized model that employs dynamic zoom-in tools for enhanced document parsing, and significantly outperforms frontier proprietary models such as GPT-5 and Gemini 2.5 Pro. 

\textbf{Spatial Reasoning Tasks.} Spatial reasoning rigorously tests a model's capacity for both granular visual perception and complex multimodal logic. We benchmark \openvl against specialized expert models fine-tuned explicitly on spatial reasoning datasets, such as RoboRefer. Despite not finetuned on this data, our model achieves the highest performance on EMbSpatial, and performs on par with the spatial-expert SpatialRGPT on the RoboSpatial, while underperformed the finetuned expert model RoboRefer-SFT marginally. Furthermore, our model significantly surpass the capabilities of GPT-5 and Gemini 2.5 Pro.

\begin{wraptable}{hr}{0.5\textwidth}
    \centering
    \footnotesize
    \renewcommand{\arraystretch}{1.1} 
    \resizebox{\linewidth}{!}{
    \begin{tabular}{l | ccc }
    \hline
    \textbf{Model} & \textbf{RefCOCO} & \textbf{RefCOCO+} & \textbf{RefCOCOg} \\
    \hline
    Gemini 1.5 Pro~\citep{gemini15report}  & 73.2  & 62.5 & 75.2 \\
    Grounding DINO~\citep{liu2023grounding}  & 90.6 & 88.2 & 86.1  \\
    \hline
    VLM-R1~\citep{shen2025vlm}           & 90.5 & 84.3 & 87.1 \\
    DeepEyes~\citep{zheng2025deepeyes}         & 89.8 & 83.6 & 86.7 \\
    OneThinker-8B~\citep{feng2025onethinkerallinonereasoningmodel}    & 92.0 & 87.0 & 89.2 \\
    \hline
    Qwen3-VL-Instruct & 89.9 & 84.5 & 86.8 \\
    Qwen3-VL GRPO    & 92.1 & 87.7 & 89.6 \\
    Qwen3-VL GDPO    & 92.2 & 87.8 & 88.9 \\
    \hline
    \rowcolor{green!5}
    \textbf{\openvl (Ours)}    & \textbf{93.4} & \textbf{88.2} & \textbf{90.4} \\
    \hline
    \end{tabular}
    }
    \vspace{-3mm}
    \caption{ Evaluation results for \colorbox{columnpurple}{\textbf{Grounding}} task. We report scores on val splits. Our model consistently outperforms previous baselines.}
    \label{tab:spatial-grounding-wrap}
    \vspace{-3mm} 
\end{wraptable}

\textbf{Grounding Tasks.} We further evaluate our model on vision-centric tasks that require formatted, high-precision numerical outputs, such as visual grounding in Table~\ref{tab:spatial-grounding-wrap}. \openvl demonstrates SOTA localization capabilities across the widely used RefCOCO, RefCOCO+, and RefCOCOg benchmarks, achieving $93.4\%$/$88.2\%$/$90.4\%$, respectively. Our model consistently outperforms previous baselines by a substantial margin, including the specialized visual grounding expert, Grounding DINO.

\textbf{Overall Performance.} In summary, \openvl exhibits balanced and robust performance across a broad spectrum of both vision-centric and reasoning-centric multimodal tasks. These extensive evaluations empirically validate the superiority of \model for multi-task RL optimization. Furthermore, they confirm that our task-level response length and entropy shaping mechanisms effectively stabilize training, accelerate convergence, and yield a highly capable multimodal generalist. We provide more details of how our model evolves its RL rewards during training, please refer to Appendix~\ref{sec:more results}. 

\subsection{Ablation Study}
\vspace{-1mm}
We conduct an ablation study to isolate the contributions of each component in \openvl, as shown in Table~\ref{tab:ablation_study}. Compared to the Qwen3-VL-8B baseline, integrating the \model objective yields the most substantial initial performance improvement. Adding task-level entropy shaping further boosts performance, particularly in reasoning-centric tasks, with more modest gains observed in saturated or out-of-distribution domains (e.g., visual grounding and spatial reasoning). Alternatively, applying the task-level length reward provides even broader improvements, outperforming entropy shaping alone and highlighting its effectiveness as a regularization directive. Ultimately, combining both shaping mechanisms with \model achieves the highest overall gains. This synergistic effect demonstrates that response length and entropy shaping are highly complementary, empirically validating the collective efficacy of our proposed methodology.

\begin{table*}[t!]
    \vspace{-10mm}
    \centering
    \footnotesize
    \setlength{\tabcolsep}{0.2em}
    \renewcommand{\arraystretch}{1.15} 
    \resizebox{\textwidth}{!}{
    \begin{tabular}{l | c c c c c c}
    \hline
    \textbf{Model} 
     & \cellcolor{columnblue}\textbf{General VQA}
     & \cellcolor{columnred}\textbf{Math VQA} 
     & \cellcolor{columnyellow}\textbf{Chart VQA}
     & \cellcolor{columnpurple}\textbf{Grounding}
     & \cellcolor{columncyan}\textbf{Document Understanding} 
     & \cellcolor{columnpeach}\textbf{Spatial Reasoning} \\ 
    \hline
    Qwen3-VL-Instruct-8B & 71.3 & 59.2 & 69.9 & 87.1 & 86.8 & 60.9 \\

    Qwen3-VL + \model    & 76.9 & 64.8 & 74.5 & 90.2 & 90.6 & 62.3 \\
    \quad + task-level entropy loss  & 77.0 & 65.1 & 75.3 & 90.4 & 90.8 & 62.8\\
    \quad + task-level length reward  & 77.4 & 65.7 &  75.4 & 90.5 & 91.1 & 63.2 \\
    \hline
    \rowcolor{green!5}
    \textbf{\openvl (Ours)}        & \textbf{77.9} & \textbf{66.2} & \textbf{76.0} & \textbf{90.7} & \textbf{91.4} & \textbf{63.6} \\
    \hline
    \end{tabular}
    }
    \caption{ Ablation study evaluating the impact of different training components across the six main domains. Scores represent the average performance within each task category.}
    \label{tab:ablation_study}
    \vspace{-5mm}
\end{table*}

\section{Related Work}

\textbf{Group Relative Policy Optimization.}
Group Relative Policy Optimization (GRPO)~\citep{guo2025deepseek}, first introduced by DeepSeek-R1, has become the de-facto Reinforcement Learning (RL) objective for enhancing the reasoning capabilities of Large Language Models (LLMs) and Multimodal LLMs (MLLMs)~\citep{zhang2025critique,dong2025agentic,yu2025dapo,xie2025logic, chen2025towards,feng2025group,liu2025understanding, zheng2025group, gao2025softadaptivepolicyoptimization}.
The success of GRPO has motivated its extension to MLLM post-training. However, the vast diversity of visual tasks exhibits significantly higher variance in reward topologies compared to standard LLM tasks, creating severe intra- and inter-task update imbalances. While recent works like EMA-GRPO~\citep{feng2025onethinkerallinonereasoningmodel} attempt to address this using task-wise moving averages of reward variance, their reliance on linear scaling fails to guarantee inter-task gradient equity and leaves the optimization vulnerable to structural pathologies. In contrast, \model resolves this by enforcing a strict Gaussian topology. This non-linear mapping mathematically caps outliers and smooths bimodal rewards, theoretically ensuring inter-task equity.

\textbf{Multimodal Reasoning.}
A growing body of literature has integrated RL into MLLMs to enable complex reasoning across diverse visual tasks~\citep{li2025star,sun2025reinforcement,feng2025video,sun2025spacevista,zhou2025reinforced,duan2025codeplot,chen2025advancing,chen2025ares,meng2025open}. A primary challenge identified in these studies is preserving robust capabilities in both fine-grained perception and multi-step reasoning~\citep{liu2025seeingbelievingprobingdisconnect,tian2025more,tu2025perceptionconsistencymultimodallargelanguage,yang2025learninglookdisentangledcurriculum}. To enhance perception, recent methods optimize auxiliary KL divergence objectives using corrupted visual inputs~\citep{wang2025perceptionawarepolicyoptimizationmultimodal, zhang2026lessrightbidirectionalperceptual}, or insert explicit visual anchors or claims into the reasoning process via proprietary models and external captioners~\citep{tian2025more, yang2025learninglookdisentangledcurriculum, tu2025perceptionconsistencymultimodallargelanguage}. However, these approaches require costly data annotations or incur substantial computational overhead, severely limiting their scalability across diverse benchmarks. We bypass these burdens by framing this challenge directly as a multi-task optimization problem. Through task-level response length and entropy shaping, our method accelerates stable convergence and effectively balances perception and reasoning capabilities.

\textbf{Optimal Transport in LLM.}
Optimal Transport (OT) is not entirely new in LLM, they are adopted in specific areas like preference alignment~\citep{melnyk2024distributionalpreferencealignmentllms,li2025optimaltransportbasedtokenweighting,na2026semanticawarewassersteinpolicyregularization,nanfack2026efficientrefusalablationllm}. Existing approaches typically leverage OT to compute semantic distances between token distributions, enforce stochastic dominance between in reward distributions, or dynamically map latent safety representations, which fundamentally treating it as a distance metric. In contrast, \model fundamentally repurposes 1D OT as a universal advantage normalization mechanism, directly addressing the extreme inter-task reward variance and heavy-tail topologies inherent in multi-domain multimodal RL.

\vspace{-3mm}
\section{Conclusion}
\vspace{-3mm}
We present \openvl, a robust, general-purpose multimodal model with multi-task reinforcement learning post-training. To address the extreme variance in reward topologies across diverse visual tasks, we propose \model to map each task's advantage distribution to converge to $\mathcal{N}(0,1)$, ensuring absolute inter-task gradient equity. Furthermore, we introduce task-level response length and entropy shaping mechanisms, effectively balancing  fine-grained perception and complex multi-step reasoning capabilities. Crucially, our method extends well beyond multimodal tasks. Since \model is inherently designed to harmonize highly divergent reward topologies, it is naturally suited for broader LLM applications that suffer from similar reward heterogeneity, such as SWE coding and GUI tasks.
We leave the large-scale expansion of this framework to future work, and we encourage the community to build upon these principles to foster more stable, scalable, and equitable RL optimization paradigms across all language and multimodal foundation models.

\section*{Acknowledgments}
This work was supported by Amazon Trainium award and compute resources, ONR grant
N00014-23-1-2780, and U.S. DARPA ANSR program FA8750-23-2-0004.



\bibliography{colm2026_conference_new}

\begin{thebibliography}{75}
\providecommand{\natexlab}[1]{#1}
\providecommand{\url}[1]{\texttt{#1}}
\expandafter\ifx\csname urlstyle\endcsname\relax
  \providecommand{\doi}[1]{doi: #1}\else
  \providecommand{\doi}{doi: \begingroup \urlstyle{rm}\Url}\fi

\bibitem[Bai et~al.(2025)Bai, Cai, Chen, Chen, Chen, Cheng, Deng, Ding, Gao, Ge, Ge, Guo, Huang, Huang, Huang, Hui, Jiang, Li, Li, Li, Li, Lin, Lin, Liu, Liu, Liu, Liu, Liu, Liu, Lu, Luo, Lv, Men, Meng, Ren, Ren, Song, Sun, Tang, Tu, Wan, Wang, Wang, Wang, Wang, Xie, Xu, Xu, Xu, Yang, Yang, Yang, Yang, Yu, Zhang, Zhang, Zhang, Zheng, Zhong, Zhou, Zhou, Zhou, Zhu, and Zhu]{qwen3vl}
Shuai Bai, Yuxuan Cai, Ruizhe Chen, Keqin Chen, Xionghui Chen, Zesen Cheng, Lianghao Deng, Wei Ding, Chang Gao, Chunjiang Ge, Wenbin Ge, Zhifang Guo, Qidong Huang, Jie Huang, Fei Huang, Binyuan Hui, Shutong Jiang, Zhaohai Li, Mingsheng Li, Mei Li, Kaixin Li, Zicheng Lin, Junyang Lin, Xuejing Liu, Jiawei Liu, Chenglong Liu, Yang Liu, Dayiheng Liu, Shixuan Liu, Dunjie Lu, Ruilin Luo, Chenxu Lv, Rui Men, Lingchen Meng, Xuancheng Ren, Xingzhang Ren, Sibo Song, Yuchong Sun, Jun Tang, Jianhong Tu, Jianqiang Wan, Peng Wang, Pengfei Wang, Qiuyue Wang, Yuxuan Wang, Tianbao Xie, Yiheng Xu, Haiyang Xu, Jin Xu, Zhibo Yang, Mingkun Yang, Jianxin Yang, An~Yang, Bowen Yu, Fei Zhang, Hang Zhang, Xi~Zhang, Bo~Zheng, Humen Zhong, Jingren Zhou, Fan Zhou, Jing Zhou, Yuanzhi Zhu, and Ke~Zhu.
\newblock Qwen3-vl technical report.
\newblock \emph{ArXiv preprint}, abs/2511.21631, 2025.
\newblock URL \url{https://arxiv.org/abs/2511.21631}.

\bibitem[Bereket \& Leskovec(2025)Bereket and Leskovec]{bereket2025uncalibrated}
Michael Bereket and Jure Leskovec.
\newblock Uncalibrated reasoning: Grpo induces overconfidence for stochastic outcomes.
\newblock \emph{ArXiv preprint}, abs/2508.11800, 2025.
\newblock URL \url{https://arxiv.org/abs/2508.11800}.

\bibitem[Chen et~al.(2024)Chen, Li, Dong, Zhang, Zang, Chen, Duan, Wang, Qiao, Lin, and Zhao]{chen2024we}
Lin Chen, Jinsong Li, Xiaoyi Dong, Pan Zhang, Yuhang Zang, Zehui Chen, Haodong Duan, Jiaqi Wang, Yu~Qiao, Dahua Lin, and Feng Zhao.
\newblock Are we on the right way for evaluating large vision-language models?
\newblock In Amir Globersons, Lester Mackey, Danielle Belgrave, Angela Fan, Ulrich Paquet, Jakub~M. Tomczak, and Cheng Zhang (eds.), \emph{Advances in Neural Information Processing Systems 38: Annual Conference on Neural Information Processing Systems 2024, NeurIPS 2024, Vancouver, BC, Canada, December 10 - 15, 2024}, 2024.
\newblock URL \url{http://papers.nips.cc/paper\_files/paper/2024/hash/2f8ee6a3d766b426d2618e555b5aeb39-Abstract-Conference.html}.

\bibitem[Chen et~al.(2025{\natexlab{a}})Chen, Qin, Liu, Peng, Guan, Wang, Hu, Zhou, Gao, and Che]{chen2025towards}
Qiguang Chen, Libo Qin, Jinhao Liu, Dengyun Peng, Jiannan Guan, Peng Wang, Mengkang Hu, Yuhang Zhou, Te~Gao, and Wanxiang Che.
\newblock Towards reasoning era: A survey of long chain-of-thought for reasoning large language models.
\newblock \emph{ArXiv preprint}, abs/2503.09567, 2025{\natexlab{a}}.
\newblock URL \url{https://arxiv.org/abs/2503.09567}.

\bibitem[Chen et~al.(2025{\natexlab{b}})Chen, Guo, Su, Li, Wu, Chen, Chen, Wang, Qu, and Cheng]{chen2025advancing}
Shuang Chen, Yue Guo, Zhaochen Su, Yafu Li, Yulun Wu, Jiacheng Chen, Jiayu Chen, Weijie Wang, Xiaoye Qu, and Yu~Cheng.
\newblock Advancing multimodal reasoning: From optimized cold start to staged reinforcement learning.
\newblock \emph{ArXiv preprint}, abs/2506.04207, 2025{\natexlab{b}}.
\newblock URL \url{https://arxiv.org/abs/2506.04207}.

\bibitem[Chen et~al.(2025{\natexlab{c}})Chen, Guo, Ye, Huang, Hu, Li, Zhang, Chen, Guo, and Peng]{chen2025ares}
Shuang Chen, Yue Guo, Yimeng Ye, Shijue Huang, Wenbo Hu, Haoxi Li, Manyuan Zhang, Jiayu Chen, Song Guo, and Nanyun Peng.
\newblock Ares: Multimodal adaptive reasoning via difficulty-aware token-level entropy shaping.
\newblock \emph{ArXiv preprint}, abs/2510.08457, 2025{\natexlab{c}}.
\newblock URL \url{https://arxiv.org/abs/2510.08457}.

\bibitem[Cheng et~al.(2024)Cheng, Yin, Fu, Guo, Yang, Kautz, Wang, and Liu]{cheng2024spatialrgpt}
An{-}Chieh Cheng, Hongxu Yin, Yang Fu, Qiushan Guo, Ruihan Yang, Jan Kautz, Xiaolong Wang, and Sifei Liu.
\newblock Spatialrgpt: Grounded spatial reasoning in vision-language models.
\newblock In Amir Globersons, Lester Mackey, Danielle Belgrave, Angela Fan, Ulrich Paquet, Jakub~M. Tomczak, and Cheng Zhang (eds.), \emph{Advances in Neural Information Processing Systems 38: Annual Conference on Neural Information Processing Systems 2024, NeurIPS 2024, Vancouver, BC, Canada, December 10 - 15, 2024}, 2024.
\newblock URL \url{http://papers.nips.cc/paper\_files/paper/2024/hash/f38cb4cf9a5eaa92b3cfa481832719c6-Abstract-Conference.html}.

\bibitem[Chu et~al.(2025)Chu, Huang, Zhang, Wei, and Wang]{chu2025gpg}
Xiangxiang Chu, Hailang Huang, Xiao Zhang, Fei Wei, and Yong Wang.
\newblock Gpg: A simple and strong reinforcement learning baseline for model reasoning.
\newblock \emph{ArXiv preprint}, abs/2504.02546, 2025.
\newblock URL \url{https://arxiv.org/abs/2504.02546}.

\bibitem[Comanici et~al.(2025)Comanici, Bieber, Schaekermann, Pasupat, Sachdeva, Dhillon, Blistein, Ram, Zhang, Rosen, et~al.]{comanici2025gemini}
Gheorghe Comanici, Eric Bieber, Mike Schaekermann, Ice Pasupat, Noveen Sachdeva, Inderjit Dhillon, Marcel Blistein, Ori Ram, Dan Zhang, Evan Rosen, et~al.
\newblock Gemini 2.5: Pushing the frontier with advanced reasoning, multimodality, long context, and next generation agentic capabilities.
\newblock \emph{ArXiv preprint}, abs/2507.06261, 2025.
\newblock URL \url{https://arxiv.org/abs/2507.06261}.

\bibitem[Deng et~al.(2025)Deng, Bansal, Yin, Peng, Wang, and Chang]{deng2025openvlthinker}
Yihe Deng, Hritik Bansal, Fan Yin, Nanyun Peng, Wei Wang, and Kai-Wei Chang.
\newblock Openvlthinker: An early exploration to complex vision-language reasoning via iterative self-improvement.
\newblock \emph{ArXiv preprint}, abs/2503.17352, 2025.
\newblock URL \url{https://arxiv.org/abs/2503.17352}.

\bibitem[Dong et~al.(2025)Dong, Mao, Ma, Bao, Chen, Wang, Chen, Du, Wang, Zhang, et~al.]{dong2025agentic}
Guanting Dong, Hangyu Mao, Kai Ma, Licheng Bao, Yifei Chen, Zhongyuan Wang, Zhongxia Chen, Jiazhen Du, Huiyang Wang, Fuzheng Zhang, et~al.
\newblock Agentic reinforced policy optimization.
\newblock \emph{ArXiv preprint}, abs/2507.19849, 2025.
\newblock URL \url{https://arxiv.org/abs/2507.19849}.

\bibitem[Du et~al.(2024)Du, Wu, Li, Huang, and Wei]{du-etal-2024-embspatial}
Mengfei Du, Binhao Wu, Zejun Li, Xuanjing Huang, and Zhongyu Wei.
\newblock {E}mb{S}patial-bench: Benchmarking spatial understanding for embodied tasks with large vision-language models.
\newblock In Lun-Wei Ku, Andre Martins, and Vivek Srikumar (eds.), \emph{Proceedings of the 62nd Annual Meeting of the Association for Computational Linguistics (Volume 2: Short Papers)}, pp.\  346--355, Bangkok, Thailand, 2024. Association for Computational Linguistics.
\newblock \doi{10.18653/v1/2024.acl-short.33}.
\newblock URL \url{https://aclanthology.org/2024.acl-short.33/}.

\bibitem[Duan et~al.(2025)Duan, Sun, Fang, Zhang, Feng, Luo, Liu, Wang, Pei, Cai, et~al.]{duan2025codeplot}
Chengqi Duan, Kaiyue Sun, Rongyao Fang, Manyuan Zhang, Yan Feng, Ying Luo, Yufang Liu, Ke~Wang, Peng Pei, Xunliang Cai, et~al.
\newblock Codeplot-cot: Mathematical visual reasoning by thinking with code-driven images.
\newblock \emph{ArXiv preprint}, abs/2510.11718, 2025.
\newblock URL \url{https://arxiv.org/abs/2510.11718}.

\bibitem[Feng et~al.(2025{\natexlab{a}})Feng, Gong, Li, Guo, Wang, Peng, Wu, Zhang, Wang, and Yue]{feng2025video}
Kaituo Feng, Kaixiong Gong, Bohao Li, Zonghao Guo, Yibing Wang, Tianshuo Peng, Junfei Wu, Xiaoying Zhang, Benyou Wang, and Xiangyu Yue.
\newblock Video-r1: Reinforcing video reasoning in mllms.
\newblock \emph{ArXiv preprint}, abs/2503.21776, 2025{\natexlab{a}}.
\newblock URL \url{https://arxiv.org/abs/2503.21776}.

\bibitem[Feng et~al.(2025{\natexlab{b}})Feng, Zhang, Li, Fan, Chen, Jiang, Zheng, Sun, Zhang, Sun, Feng, Pei, Cai, and Yue]{feng2025onethinkerallinonereasoningmodel}
Kaituo Feng, Manyuan Zhang, Hongyu Li, Kaixuan Fan, Shuang Chen, Yilei Jiang, Dian Zheng, Peiwen Sun, Yiyuan Zhang, Haoze Sun, Yan Feng, Peng Pei, Xunliang Cai, and Xiangyu Yue.
\newblock Onethinker: All-in-one reasoning model for image and video, 2025{\natexlab{b}}.
\newblock URL \url{https://arxiv.org/abs/2512.03043}.

\bibitem[Feng et~al.(2025{\natexlab{c}})Feng, Xue, Liu, and An]{feng2025group}
Lang Feng, Zhenghai Xue, Tingcong Liu, and Bo~An.
\newblock Group-in-group policy optimization for llm agent training.
\newblock \emph{ArXiv preprint}, abs/2505.10978, 2025{\natexlab{c}}.
\newblock URL \url{https://arxiv.org/abs/2505.10978}.

\bibitem[Foutter et~al.(2024)Foutter, Gammelli, Kruger, Foss, Bhoj, Guffanti, D'Amico, and Pavone]{foutter2025spacellavavisionlanguagemodeladapted}
Matthew Foutter, Daniele Gammelli, Justin Kruger, Ethan Foss, Praneet Bhoj, Tommaso Guffanti, Simone D'Amico, and Marco Pavone.
\newblock Space-llava: a vision-language model adapted to extraterrestrial applications, 2024.
\newblock URL \url{https://arxiv.org/abs/2408.05924}.

\bibitem[Gao et~al.(2025)Gao, Zheng, Chen, Dang, Liu, Yu, Yang, Bai, Zhou, and Lin]{gao2025softadaptivepolicyoptimization}
Chang Gao, Chujie Zheng, Xiong-Hui Chen, Kai Dang, Shixuan Liu, Bowen Yu, An~Yang, Shuai Bai, Jingren Zhou, and Junyang Lin.
\newblock Soft adaptive policy optimization, 2025.
\newblock URL \url{https://arxiv.org/abs/2511.20347}.

\bibitem[Gemini~Team(2024)]{gemini15report}
Google Gemini~Team.
\newblock Gemini 1.5: Unlocking multimodal understanding across millions of tokens of context.
\newblock \emph{ArXiv preprint}, abs/2403.05530, 2024.
\newblock URL \url{https://arxiv.org/abs/2403.05530}.

\bibitem[Guo et~al.(2025)Guo, Yang, Zhang, Song, Zhang, Xu, Zhu, Ma, Wang, Bi, et~al.]{guo2025deepseek}
Daya Guo, Dejian Yang, Haowei Zhang, Junxiao Song, Ruoyu Zhang, Runxin Xu, Qihao Zhu, Shirong Ma, Peiyi Wang, Xiao Bi, et~al.
\newblock Deepseek-r1: Incentivizing reasoning capability in llms via reinforcement learning.
\newblock \emph{ArXiv preprint}, abs/2501.12948, 2025.
\newblock URL \url{https://arxiv.org/abs/2501.12948}.

\bibitem[Hong et~al.(2025)Hong, Zhao, Zhu, Lu, Xu, and Yu]{hong2025deepeyesv2}
Jack Hong, Chenxiao Zhao, ChengLin Zhu, Weiheng Lu, Guohai Xu, and Xing Yu.
\newblock Deepeyesv2: Toward agentic multimodal model.
\newblock \emph{ArXiv preprint}, abs/2511.05271, 2025.
\newblock URL \url{https://arxiv.org/abs/2511.05271}.

\bibitem[Hu et~al.(2025)Hu, Lin, Long, Ran, Jiang, Wang, Zhu, Xu, Wang, and Pang]{hu2025g2vlmgeometrygroundedvision}
Wenbo Hu, Jingli Lin, Yilin Long, Yunlong Ran, Lihan Jiang, Yifan Wang, Chenming Zhu, Runsen Xu, Tai Wang, and Jiangmiao Pang.
\newblock G$^2$vlm: Geometry grounded vision language model with unified 3d reconstruction and spatial reasoning.
\newblock \emph{ArXiv preprint}, abs/2511.21688, 2025.
\newblock URL \url{https://arxiv.org/abs/2511.21688}.

\bibitem[Huang et~al.(2025)Huang, Zhang, Fang, Liang, Rong, Yao, Wan, Liang, He, Li, et~al.]{huang2025mapo}
Wenke Huang, Quan Zhang, Yiyang Fang, Jian Liang, Xuankun Rong, Huanjin Yao, Guancheng Wan, Ke~Liang, Wenwen He, Mingjun Li, et~al.
\newblock Mapo: Mixed advantage policy optimization.
\newblock \emph{ArXiv preprint}, abs/2509.18849, 2025.
\newblock URL \url{https://arxiv.org/abs/2509.18849}.

\bibitem[Hurst et~al.(2024)Hurst, Lerer, Goucher, Perelman, Ramesh, Clark, Ostrow, Welihinda, Hayes, Radford, et~al.]{hurst2024gpt}
Aaron Hurst, Adam Lerer, Adam~P Goucher, Adam Perelman, Aditya Ramesh, Aidan Clark, AJ~Ostrow, Akila Welihinda, Alan Hayes, Alec Radford, et~al.
\newblock Gpt-4o system card.
\newblock \emph{ArXiv preprint}, abs/2410.21276, 2024.
\newblock URL \url{https://arxiv.org/abs/2410.21276}.

\bibitem[Kazemzadeh et~al.(2014)Kazemzadeh, Ordonez, Matten, and Berg]{kazemzadeh2014referitgame}
Sahar Kazemzadeh, Vicente Ordonez, Mark Matten, and Tamara Berg.
\newblock {R}efer{I}t{G}ame: Referring to objects in photographs of natural scenes.
\newblock In Alessandro Moschitti, Bo~Pang, and Walter Daelemans (eds.), \emph{Proceedings of the 2014 Conference on Empirical Methods in Natural Language Processing ({EMNLP})}, pp.\  787--798, Doha, Qatar, 2014. Association for Computational Linguistics.
\newblock \doi{10.3115/v1/D14-1086}.
\newblock URL \url{https://aclanthology.org/D14-1086}.

\bibitem[Kembhavi et~al.(2016)Kembhavi, Salvato, Kolve, Seo, Hajishirzi, and Farhadi]{kembhavi2016diagram}
Aniruddha Kembhavi, Mike Salvato, Eric Kolve, Minjoon Seo, Hannaneh Hajishirzi, and Ali Farhadi.
\newblock A diagram is worth a dozen images.
\newblock In \emph{European conference on computer vision}, pp.\  235--251. Springer, 2016.

\bibitem[Li et~al.(2025{\natexlab{a}})Li, Huzhang, Zhang, Wang, and Zeng]{li2025optimaltransportbasedtokenweighting}
Meng Li, Guangda Huzhang, Haibo Zhang, Xiting Wang, and Anxiang Zeng.
\newblock Optimal transport-based token weighting scheme for enhanced preference optimization, 2025{\natexlab{a}}.
\newblock URL \url{https://arxiv.org/abs/2505.18720}.

\bibitem[Li et~al.(2025{\natexlab{b}})Li, Ma, Li, Li, Rong, Xu, Zhang, Zhao, and Huang]{li2025star}
Zongzhao Li, Zongyang Ma, Mingze Li, Songyou Li, Yu~Rong, Tingyang Xu, Ziqi Zhang, Deli Zhao, and Wenbing Huang.
\newblock Star-r1: Spatial transformation reasoning by reinforcing multimodal llms.
\newblock \emph{ArXiv preprint}, abs/2505.15804, 2025{\natexlab{b}}.
\newblock URL \url{https://arxiv.org/abs/2505.15804}.

\bibitem[Liu et~al.(2023)Liu, Zeng, Ren, Li, Zhang, Yang, Li, Yang, Su, Zhu, et~al.]{liu2023grounding}
Shilong Liu, Zhaoyang Zeng, Tianhe Ren, Feng Li, Hao Zhang, Jie Yang, Chunyuan Li, Jianwei Yang, Hang Su, Jun Zhu, et~al.
\newblock Grounding dino: Marrying dino with grounded pre-training for open-set object detection.
\newblock \emph{ArXiv preprint}, abs/2303.05499, 2023.
\newblock URL \url{https://arxiv.org/abs/2303.05499}.

\bibitem[Liu et~al.(2024{\natexlab{a}})Liu, Duan, Zhang, Li, Zhang, Zhao, Yuan, Wang, He, Liu, et~al.]{liu2024mmbench}
Yuan Liu, Haodong Duan, Yuanhan Zhang, Bo~Li, Songyang Zhang, Wangbo Zhao, Yike Yuan, Jiaqi Wang, Conghui He, Ziwei Liu, et~al.
\newblock Mmbench: Is your multi-modal model an all-around player?
\newblock In \emph{European conference on computer vision}, pp.\  216--233. Springer, 2024{\natexlab{a}}.

\bibitem[Liu et~al.(2024{\natexlab{b}})Liu, Li, Huang, Yang, Yu, Li, Yin, Liu, Jin, and Bai]{OCRBench}
Yuliang Liu, Zhang Li, Mingxin Huang, Biao Yang, Wenwen Yu, Chunyuan Li, Xu-Cheng Yin, Cheng-Lin Liu, Lianwen Jin, and Xiang Bai.
\newblock Ocrbench: on the hidden mystery of ocr in large multimodal models.
\newblock \emph{Science China Information Sciences}, 67\penalty0 (12), December 2024{\natexlab{b}}.
\newblock ISSN 1869-1919.
\newblock \doi{10.1007/s11432-024-4235-6}.
\newblock URL \url{http://dx.doi.org/10.1007/s11432-024-4235-6}.

\bibitem[Liu et~al.(2025{\natexlab{a}})Liu, Chen, Liu, Luo, Tang, Wang, Zeng, Dai, Shi, Wei, Dumoulin, and Tong]{liu2025seeingbelievingprobingdisconnect}
Zhining Liu, Ziyi Chen, Hui Liu, Chen Luo, Xianfeng Tang, Suhang Wang, Joy Zeng, Zhenwei Dai, Zhan Shi, Tianxin Wei, Benoit Dumoulin, and Hanghang Tong.
\newblock Seeing but not believing: Probing the disconnect between visual attention and answer correctness in vlms, 2025{\natexlab{a}}.
\newblock URL \url{https://arxiv.org/abs/2510.17771}.

\bibitem[Liu et~al.(2025{\natexlab{b}})Liu, Chen, Li, Qi, Pang, Du, Lee, and Lin]{liu2025understanding}
Zichen Liu, Changyu Chen, Wenjun Li, Penghui Qi, Tianyu Pang, Chao Du, Wee~Sun Lee, and Min Lin.
\newblock Understanding r1-zero-like training: A critical perspective.
\newblock \emph{ArXiv preprint}, abs/2503.20783, 2025{\natexlab{b}}.
\newblock URL \url{https://arxiv.org/abs/2503.20783}.

\bibitem[Liu et~al.(2025{\natexlab{c}})Liu, Sun, Zang, Dong, Cao, Duan, Lin, and Wang]{liu2025visual}
Ziyu Liu, Zeyi Sun, Yuhang Zang, Xiaoyi Dong, Yuhang Cao, Haodong Duan, Dahua Lin, and Jiaqi Wang.
\newblock Visual-rft: Visual reinforcement fine-tuning.
\newblock \emph{ArXiv preprint}, abs/2503.01785, 2025{\natexlab{c}}.
\newblock URL \url{https://arxiv.org/abs/2503.01785}.

\bibitem[Lu et~al.(2024)Lu, Bansal, Xia, Liu, Li, Hajishirzi, Cheng, Chang, Galley, and Gao]{lu2023mathvista}
Pan Lu, Hritik Bansal, Tony Xia, Jiacheng Liu, Chunyuan Li, Hannaneh Hajishirzi, Hao Cheng, Kai{-}Wei Chang, Michel Galley, and Jianfeng Gao.
\newblock Mathvista: Evaluating mathematical reasoning of foundation models in visual contexts.
\newblock In \emph{The Twelfth International Conference on Learning Representations, {ICLR} 2024, Vienna, Austria, May 7-11, 2024}. OpenReview.net, 2024.
\newblock URL \url{https://openreview.net/forum?id=KUNzEQMWU7}.

\bibitem[Masry et~al.(2022)Masry, Do, Tan, Joty, and Hoque]{masry-etal-2022-chartqa}
Ahmed Masry, Xuan~Long Do, Jia~Qing Tan, Shafiq Joty, and Enamul Hoque.
\newblock {C}hart{QA}: A benchmark for question answering about charts with visual and logical reasoning.
\newblock In Smaranda Muresan, Preslav Nakov, and Aline Villavicencio (eds.), \emph{Findings of the Association for Computational Linguistics: ACL 2022}, pp.\  2263--2279, Dublin, Ireland, 2022. Association for Computational Linguistics.
\newblock \doi{10.18653/v1/2022.findings-acl.177}.
\newblock URL \url{https://aclanthology.org/2022.findings-acl.177}.

\bibitem[Mathew et~al.(2021)Mathew, Karatzas, and Jawahar]{mathew2021docvqa}
Minesh Mathew, Dimosthenis Karatzas, and CV~Jawahar.
\newblock Docvqa: A dataset for vqa on document images.
\newblock In \emph{Proceedings of the IEEE/CVF Winter Conference on Applications of Computer Vision}, pp.\  2199--2208, 2021.

\bibitem[Mathew et~al.(2022)Mathew, Bagal, Tito, Karatzas, Valveny, and Jawahar]{mathew2022infographicvqa}
Minesh Mathew, Viraj Bagal, Rub{\`e}n~P{\'e}rez Tito, Dimosthenis Karatzas, Ernest Valveny, and CV~Jawahar.
\newblock Infographicvqa.
\newblock In \emph{Proceedings of the IEEE/CVF Winter Conference on Applications of Computer Vision}, pp.\  2821--2831, 2022.

\bibitem[Melnyk et~al.(2024)Melnyk, Mroueh, Belgodere, Rigotti, Nitsure, Yurochkin, Greenewald, Navratil, and Ross]{melnyk2024distributionalpreferencealignmentllms}
Igor Melnyk, Youssef Mroueh, Brian Belgodere, Mattia Rigotti, Apoorva Nitsure, Mikhail Yurochkin, Kristjan Greenewald, Jiri Navratil, and Jerret Ross.
\newblock Distributional preference alignment of llms via optimal transport, 2024.
\newblock URL \url{https://arxiv.org/abs/2406.05882}.

\bibitem[Meng et~al.(2025{\natexlab{a}})Meng, Du, Liu, Zhou, Lu, Fu, Han, Shi, Wang, He, et~al.]{meng2025mm}
Fanqing Meng, Lingxiao Du, Zongkai Liu, Zhixiang Zhou, Quanfeng Lu, Daocheng Fu, Tiancheng Han, Botian Shi, Wenhai Wang, Junjun He, et~al.
\newblock Mm-eureka: Exploring the frontiers of multimodal reasoning with rule-based reinforcement learning.
\newblock \emph{ArXiv preprint}, abs/2503.07365, 2025{\natexlab{a}}.
\newblock URL \url{https://arxiv.org/abs/2503.07365}.

\bibitem[Meng et~al.(2025{\natexlab{b}})Meng, Li, Wang, Tan, Zhang, Kong, Tong, Wang, Teng, Wang, et~al.]{meng2025open}
Jiahao Meng, Xiangtai Li, Haochen Wang, Yue Tan, Tao Zhang, Lingdong Kong, Yunhai Tong, Anran Wang, Zhiyang Teng, Yujing Wang, et~al.
\newblock Open-o3 video: Grounded video reasoning with explicit spatio-temporal evidence.
\newblock \emph{ArXiv preprint}, abs/2510.20579, 2025{\natexlab{b}}.
\newblock URL \url{https://arxiv.org/abs/2510.20579}.

\bibitem[Na et~al.(2026)Na, Na, Kim, Jo, Bae, Kang, and Moon]{na2026semanticawarewassersteinpolicyregularization}
Byeonghu Na, Hyungho Na, Yeongmin Kim, Suhyeon Jo, HeeSun Bae, Mina Kang, and Il-Chul Moon.
\newblock Semantic-aware wasserstein policy regularization for large language model alignment, 2026.
\newblock URL \url{https://arxiv.org/abs/2602.01685}.

\bibitem[Nanfack et~al.(2026)Nanfack, Belilovsky, and Dohmatob]{nanfack2026efficientrefusalablationllm}
Geraldin Nanfack, Eugene Belilovsky, and Elvis Dohmatob.
\newblock Efficient refusal ablation in llm through optimal transport, 2026.
\newblock URL \url{https://arxiv.org/abs/2603.04355}.

\bibitem[Seed(2026)]{seed2026seed18modelcardgeneralized}
Bytedance Seed.
\newblock Seed1.8 model card: Towards generalized real-world agency, 2026.
\newblock URL \url{https://arxiv.org/abs/2603.20633}.

\bibitem[Shen et~al.(2025)Shen, Liu, Li, Fang, Ma, Liao, Shen, Zhang, Zhao, Zhang, et~al.]{shen2025vlm}
Haozhan Shen, Peng Liu, Jingcheng Li, Chunxin Fang, Yibo Ma, Jiajia Liao, Qiaoli Shen, Zilun Zhang, Kangjia Zhao, Qianqian Zhang, et~al.
\newblock Vlm-r1: A stable and generalizable r1-style large vision-language model.
\newblock \emph{ArXiv preprint}, abs/2504.07615, 2025.
\newblock URL \url{https://arxiv.org/abs/2504.07615}.

\bibitem[Singh et~al.(2026)Singh, Fry, Perelman, Tart, Ganesh, El-Kishky, McLaughlin, Low, Ostrow, Ananthram, Nathan, Luo, Helyar, Madry, Efremov, Spyra, Baker-Whitcomb, Beutel, Karpenko, Makelov, Neitz, Wei, Barr, Kirchmeyer, et~al.]{singh2025openaigpt5card}
Aaditya Singh, Adam Fry, Adam Perelman, Adam Tart, Adi Ganesh, Ahmed El-Kishky, Aidan McLaughlin, Aiden Low, AJ~Ostrow, Akhila Ananthram, Akshay Nathan, Alan Luo, Alec Helyar, Aleksander Madry, Aleksandr Efremov, Aleksandra Spyra, Alex Baker-Whitcomb, Alex Beutel, Alex Karpenko, Alex Makelov, Alex Neitz, Alex Wei, Alexandra Barr, Alexandre Kirchmeyer, et~al.
\newblock Openai gpt-5 system card, 2026.
\newblock URL \url{https://arxiv.org/abs/2601.03267}.

\bibitem[Song et~al.(2025)Song, Blukis, Tremblay, Tyree, Su, and Birchfield]{song2025robospatial}
Chan~Hee Song, Valts Blukis, Jonathan Tremblay, Stephen Tyree, Yu~Su, and Stan Birchfield.
\newblock {RoboSpatial}: Teaching spatial understanding to {2D} and {3D} vision-language models for robotics.
\newblock In \emph{Proceedings of the IEEE/CVF Conference on Computer Vision and Pattern Recognition (CVPR)}, 2025.
\newblock Oral Presentation.

\bibitem[Sun et~al.(2025{\natexlab{a}})Sun, Wu, Xia, Luo, Zhao, Qin, Lv, Zhang, Chang, and Wang]{sun2025reinforcement}
Haoyuan Sun, Jiaqi Wu, Bo~Xia, Yifu Luo, Yifei Zhao, Kai Qin, Xufei Lv, Tiantian Zhang, Yongzhe Chang, and Xueqian Wang.
\newblock Reinforcement fine-tuning powers reasoning capability of multimodal large language models.
\newblock \emph{ArXiv preprint}, abs/2505.18536, 2025{\natexlab{a}}.
\newblock URL \url{https://arxiv.org/abs/2505.18536}.

\bibitem[Sun et~al.(2025{\natexlab{b}})Sun, Lang, Wu, Ding, Feng, Liu, Ye, Liu, Liu, Wang, et~al.]{sun2025spacevista}
Peiwen Sun, Shiqiang Lang, Dongming Wu, Yi~Ding, Kaituo Feng, Huadai Liu, Zhen Ye, Rui Liu, Yun-Hui Liu, Jianan Wang, et~al.
\newblock Spacevista: All-scale visual spatial reasoning from mm to km.
\newblock \emph{ArXiv preprint}, abs/2510.09606, 2025{\natexlab{b}}.
\newblock URL \url{https://arxiv.org/abs/2510.09606}.

\bibitem[Team et~al.(2025)Team, Cao, Tan, Ji, Chen, Lin, Li, Cao, Wang, Zhou, Han, Tang, Xu, Guo, Lyu, Xu, Shi, Du, Chi, Zhao, Hao, Zhao, Zhang, Rong, Lyu, Cai, Fu, Chen, Zhang, Zhang, Zhang, Liu, Feng, Wang, Liu, Jiao, Lyu, Chen, He, Ao, Sun, He, Zheng, Yang, Shi, Xie, Zhang, Nie, Men, Lin, Wang, Huang, and Zhang]{baairobobrainteam2025robobrain20technicalreport}
BAAI~RoboBrain Team, Mingyu Cao, Huajie Tan, Yuheng Ji, Xiansheng Chen, Minglan Lin, Zhiyu Li, Zhou Cao, Pengwei Wang, Enshen Zhou, Yi~Han, Yingbo Tang, Xiangqi Xu, Wei Guo, Yaoxu Lyu, Yijie Xu, Jiayu Shi, Mengfei Du, Cheng Chi, Mengdi Zhao, Xiaoshuai Hao, Junkai Zhao, Xiaojie Zhang, Shanyu Rong, Huaihai Lyu, Zhengliang Cai, Yankai Fu, Ning Chen, Bolun Zhang, Lingfeng Zhang, Shuyi Zhang, Dong Liu, Xi~Feng, Songjing Wang, Xiaodan Liu, Yance Jiao, Mengsi Lyu, Zhuo Chen, Chenrui He, Yulong Ao, Xue Sun, Zheqi He, Jingshu Zheng, Xi~Yang, Donghai Shi, Kunchang Xie, Bochao Zhang, Shaokai Nie, Chunlei Men, Yonghua Lin, Zhongyuan Wang, Tiejun Huang, and Shanghang Zhang.
\newblock Robobrain 2.0 technical report, 2025.
\newblock URL \url{https://arxiv.org/abs/2507.02029}.

\bibitem[Team(2026)]{kimiteam2026kimik25visualagentic}
Kimi Team.
\newblock Kimi k2.5: Visual agentic intelligence, 2026.
\newblock URL \url{https://arxiv.org/abs/2602.02276}.

\bibitem[Tian et~al.(2025)Tian, Zou, Yang, He, Waschkowski, Wesemann, Tu, and Zhang]{tian2025more}
Xinyu Tian, Shu Zou, Zhaoyuan Yang, Mengqi He, Fabian Waschkowski, Lukas Wesemann, Peter Tu, and Jing Zhang.
\newblock More thought, less accuracy? on the dual nature of reasoning in vision-language models.
\newblock \emph{ArXiv preprint}, abs/2509.25848, 2025.
\newblock URL \url{https://arxiv.org/abs/2509.25848}.

\bibitem[Tu et~al.(2025)Tu, Zhang, Sun, Fu, Li, Lan, Jiang, Wang, and Zhao]{tu2025perceptionconsistencymultimodallargelanguage}
Songjun Tu, Qichao Zhang, Jingbo Sun, Yuqian Fu, Linjing Li, Xiangyuan Lan, Dongmei Jiang, Yaowei Wang, and Dongbin Zhao.
\newblock Perception-consistency multimodal large language models reasoning via caption-regularized policy optimization, 2025.
\newblock URL \url{https://arxiv.org/abs/2509.21854}.

\bibitem[Wang et~al.(2025{\natexlab{a}})Wang, Fan, Yang, Hu, Karimi, Yao, and Yang]{wang2025regiondocr1}
Chao Wang, Hehe Fan, Huichen Yang, Zhengdong Hu, Sarvnaz Karimi, Lina Yao, and Yi~Yang.
\newblock Regiondoc-r1: Reinforcing semantic layout-aware learning for document understanding, 2025{\natexlab{a}}.
\newblock URL \url{https://openreview.net/forum?id=pfHm4YJTzC}.

\bibitem[Wang et~al.(2025{\natexlab{b}})Wang, Qu, Huang, Chu, Lin, and Chen]{wang2025vl}
Haozhe Wang, Chao Qu, Zuming Huang, Wei Chu, Fangzhen Lin, and Wenhu Chen.
\newblock Vl-rethinker: Incentivizing self-reflection of vision-language models with reinforcement learning.
\newblock \emph{ArXiv preprint}, abs/2504.08837, 2025{\natexlab{b}}.
\newblock URL \url{https://arxiv.org/abs/2504.08837}.

\bibitem[Wang et~al.(2024{\natexlab{a}})Wang, Pan, Shi, Lu, Ren, Zhou, Zhan, and Li]{wang2024measuring}
Ke~Wang, Junting Pan, Weikang Shi, Zimu Lu, Houxing Ren, Aojun Zhou, Mingjie Zhan, and Hongsheng Li.
\newblock Measuring multimodal mathematical reasoning with math-vision dataset.
\newblock In Amir Globersons, Lester Mackey, Danielle Belgrave, Angela Fan, Ulrich Paquet, Jakub~M. Tomczak, and Cheng Zhang (eds.), \emph{Advances in Neural Information Processing Systems 38: Annual Conference on Neural Information Processing Systems 2024, NeurIPS 2024, Vancouver, BC, Canada, December 10 - 15, 2024}, 2024{\natexlab{a}}.
\newblock URL \url{http://papers.nips.cc/paper\_files/paper/2024/hash/ad0edc7d5fa1a783f063646968b7315b-Abstract-Datasets\_and\_Benchmarks\_Track.html}.

\bibitem[Wang et~al.(2025{\natexlab{c}})Wang, Liu, Zhou, Shi, Lin, Chen, Li, Wan, and Zhao]{wang2026visionzeroscalablevlmselfimprovement}
Qinsi Wang, Bo~Liu, Tianyi Zhou, Jing Shi, Yueqian Lin, Yiran Chen, Hai~Helen Li, Kun Wan, and Wentian Zhao.
\newblock Vision-zero: Scalable vlm self-improvement via strategic gamified self-play, 2025{\natexlab{c}}.
\newblock URL \url{https://arxiv.org/abs/2509.25541}.

\bibitem[Wang et~al.(2025{\natexlab{d}})Wang, Guo, Stoica, Xu, Wang, Ha, Chen, Chen, Yan, Huang, and Ji]{wang2025perceptionawarepolicyoptimizationmultimodal}
Zhenhailong Wang, Xuehang Guo, Sofia Stoica, Haiyang Xu, Hongru Wang, Hyeonjeong Ha, Xiusi Chen, Yangyi Chen, Ming Yan, Fei Huang, and Heng Ji.
\newblock Perception-aware policy optimization for multimodal reasoning, 2025{\natexlab{d}}.
\newblock URL \url{https://arxiv.org/abs/2507.06448}.

\bibitem[Wang et~al.(2024{\natexlab{b}})Wang, Xia, He, Chen, Liu, Zhu, Liang, Wu, Liu, Malladi, Chevalier, Arora, and Chen]{wang2024charxiv}
Zirui Wang, Mengzhou Xia, Luxi He, Howard Chen, Yitao Liu, Richard Zhu, Kaiqu Liang, Xindi Wu, Haotian Liu, Sadhika Malladi, Alexis Chevalier, Sanjeev Arora, and Danqi Chen.
\newblock Charxiv: Charting gaps in realistic chart understanding in multimodal llms.
\newblock In Amir Globersons, Lester Mackey, Danielle Belgrave, Angela Fan, Ulrich Paquet, Jakub~M. Tomczak, and Cheng Zhang (eds.), \emph{Advances in Neural Information Processing Systems 38: Annual Conference on Neural Information Processing Systems 2024, NeurIPS 2024, Vancouver, BC, Canada, December 10 - 15, 2024}, 2024{\natexlab{b}}.
\newblock URL \url{http://papers.nips.cc/paper\_files/paper/2024/hash/cdf6f8e9fd9aeaf79b6024caec24f15b-Abstract-Datasets\_and\_Benchmarks\_Track.html}.

\bibitem[Wei et~al.(2025)Wei, Zhao, Sun, Lin, Yin, Hu, Zhang, Yu, Lv, Weng, Wang, Han, Peng, Han, Ge, Zhang, Jiang, and Patel]{ovrwei2025openvisionreasonertransferring}
Yana Wei, Liang Zhao, Jianjian Sun, Kangheng Lin, Jisheng Yin, Jingcheng Hu, Yinmin Zhang, En~Yu, Haoran Lv, Zejia Weng, Jia Wang, Chunrui Han, Yuang Peng, Qi~Han, Zheng Ge, Xiangyu Zhang, Daxin Jiang, and Vishal~M. Patel.
\newblock Open vision reasoner: Transferring linguistic cognitive behavior for visual reasoning, 2025.
\newblock URL \url{https://arxiv.org/abs/2507.05255}.

\bibitem[Xie et~al.(2025)Xie, Gao, Ren, Luo, Hong, Dai, Zhou, Qiu, Wu, and Luo]{xie2025logic}
Tian Xie, Zitian Gao, Qingnan Ren, Haoming Luo, Yuqian Hong, Bryan Dai, Joey Zhou, Kai Qiu, Zhirong Wu, and Chong Luo.
\newblock Logic-rl: Unleashing llm reasoning with rule-based reinforcement learning.
\newblock \emph{ArXiv preprint}, abs/2502.14768, 2025.
\newblock URL \url{https://arxiv.org/abs/2502.14768}.

\bibitem[Yang et~al.(2025{\natexlab{a}})Yang, Li, Lai, Yu, Zhao, and Jia]{yang2025visionthink}
Senqiao Yang, Junyi Li, Xin Lai, Bei Yu, Hengshuang Zhao, and Jiaya Jia.
\newblock Visionthink: Smart and efficient vision language model via reinforcement learning.
\newblock \emph{ArXiv preprint}, abs/2507.13348, 2025{\natexlab{a}}.
\newblock URL \url{https://arxiv.org/abs/2507.13348}.

\bibitem[Yang et~al.(2025{\natexlab{b}})Yang, Gao, Qiu, Liu, Shi, Zeng, Liao, and Ma]{yang2025learninglookdisentangledcurriculum}
Siqi Yang, Zilve Gao, Haibo Qiu, Fanfan Liu, Peng Shi, Zhixiong Zeng, Qingmin Liao, and Lin Ma.
\newblock Learning when to look: A disentangled curriculum for strategic perception in multimodal reasoning, 2025{\natexlab{b}}.
\newblock URL \url{https://arxiv.org/abs/2512.17227}.

\bibitem[Yu et~al.(2016)Yu, Poirson, Yang, Berg, and Berg]{yu2016modeling}
Licheng Yu, Patrick Poirson, Shan Yang, Alexander~C Berg, and Tamara~L Berg.
\newblock Modeling context in referring expressions.
\newblock In \emph{European conference on computer vision}, pp.\  69--85. Springer, 2016.

\bibitem[Yu et~al.(2025{\natexlab{a}})Yu, Zhang, Zhu, Yuan, Zuo, Yue, Dai, Fan, Liu, Liu, et~al.]{yu2025dapo}
Qiying Yu, Zheng Zhang, Ruofei Zhu, Yufeng Yuan, Xiaochen Zuo, Yu~Yue, Weinan Dai, Tiantian Fan, Gaohong Liu, Lingjun Liu, et~al.
\newblock Dapo: An open-source llm reinforcement learning system at scale.
\newblock \emph{ArXiv preprint}, abs/2503.14476, 2025{\natexlab{a}}.
\newblock URL \url{https://arxiv.org/abs/2503.14476}.

\bibitem[Yu et~al.(2025{\natexlab{b}})Yu, Yang, Liu, and Bai]{yu2025docthinkerexplainablemultimodallarge}
Wenwen Yu, Zhibo Yang, Yuliang Liu, and Xiang Bai.
\newblock Docthinker: Explainable multimodal large language models with rule-based reinforcement learning for document understanding, 2025{\natexlab{b}}.
\newblock URL \url{https://arxiv.org/abs/2508.08589}.

\bibitem[Yue et~al.(2024)Yue, Ni, Zheng, Zhang, Liu, Zhang, Stevens, Jiang, Ren, Sun, Wei, Yu, Yuan, Sun, Yin, Zheng, Yang, Liu, Huang, Sun, Su, and Chen]{yue2024mmmu}
Xiang Yue, Yuansheng Ni, Tianyu Zheng, Kai Zhang, Ruoqi Liu, Ge~Zhang, Samuel Stevens, Dongfu Jiang, Weiming Ren, Yuxuan Sun, Cong Wei, Botao Yu, Ruibin Yuan, Renliang Sun, Ming Yin, Boyuan Zheng, Zhenzhu Yang, Yibo Liu, Wenhao Huang, Huan Sun, Yu~Su, and Wenhu Chen.
\newblock {MMMU:} {A} massive multi-discipline multimodal understanding and reasoning benchmark for expert {AGI}.
\newblock In \emph{{IEEE/CVF} Conference on Computer Vision and Pattern Recognition, {CVPR} 2024, Seattle, WA, USA, June 16-22, 2024}, pp.\  9556--9567. {IEEE}, 2024.
\newblock \doi{10.1109/CVPR52733.2024.00913}.
\newblock URL \url{https://doi.org/10.1109/CVPR52733.2024.00913}.

\bibitem[Zha et~al.(2025)Zha, Zhou, Wu, Wang, Feng, Xu, Hao, Liu, Xing, and Hu]{zha2025visiong1generalvisionlanguage}
Yuheng Zha, Kun Zhou, Yujia Wu, Yushu Wang, Jie Feng, Zhi Xu, Shibo Hao, Zhengzhong Liu, Eric~P. Xing, and Zhiting Hu.
\newblock Vision-g1: Towards general vision language reasoning with multi-domain data curation, 2025.
\newblock URL \url{https://arxiv.org/abs/2508.12680}.

\bibitem[Zhang et~al.(2024)Zhang, Jiang, Zhang, Lin, Guo, Qiu, Zhou, Lu, Chang, Qiao, et~al.]{zhang2024mathverse}
Renrui Zhang, Dongzhi Jiang, Yichi Zhang, Haokun Lin, Ziyu Guo, Pengshuo Qiu, Aojun Zhou, Pan Lu, Kai-Wei Chang, Yu~Qiao, et~al.
\newblock Mathverse: Does your multi-modal llm truly see the diagrams in visual math problems?
\newblock In \emph{European Conference on Computer Vision}, pp.\  169--186. Springer, 2024.

\bibitem[Zhang et~al.(2025{\natexlab{a}})Zhang, Zhang, Fu, Song, Bian, Yang, and Wang]{zhang2026lessrightbidirectionalperceptual}
Shuoshuo Zhang, Yizhen Zhang, Jingjing Fu, Lei Song, Jiang Bian, Yujiu Yang, and Rui Wang.
\newblock See less, see right: Bi-directional perceptual shaping for multimodal reasoning, 2025{\natexlab{a}}.
\newblock URL \url{https://arxiv.org/abs/2512.22120}.

\bibitem[Zhang et~al.(2025{\natexlab{b}})Zhang, Sun, Zhang, Feng, Lu, Yang, and Meng]{zhang2025critique}
Xiaoying Zhang, Hao Sun, Yipeng Zhang, Kaituo Feng, Chaochao Lu, Chao Yang, and Helen Meng.
\newblock Critique-grpo: Advancing llm reasoning with natural language and numerical feedback.
\newblock \emph{ArXiv preprint}, abs/2506.03106, 2025{\natexlab{b}}.
\newblock URL \url{https://arxiv.org/abs/2506.03106}.

\bibitem[Zheng et~al.(2025{\natexlab{a}})Zheng, Liu, Li, Chen, Yu, Gao, Dang, Liu, Men, Yang, et~al.]{zheng2025group}
Chujie Zheng, Shixuan Liu, Mingze Li, Xiong-Hui Chen, Bowen Yu, Chang Gao, Kai Dang, Yuqiong Liu, Rui Men, An~Yang, et~al.
\newblock Group sequence policy optimization.
\newblock \emph{ArXiv preprint}, abs/2507.18071, 2025{\natexlab{a}}.
\newblock URL \url{https://arxiv.org/abs/2507.18071}.

\bibitem[Zheng et~al.(2025{\natexlab{b}})Zheng, Yang, Hong, Zhao, Xu, Yang, Shen, and Yu]{zheng2025deepeyes}
Ziwei Zheng, Michael Yang, Jack Hong, Chenxiao Zhao, Guohai Xu, Le~Yang, Chao Shen, and Xing Yu.
\newblock Deepeyes: Incentivizing" thinking with images" via reinforcement learning.
\newblock \emph{ArXiv preprint}, abs/2505.14362, 2025{\natexlab{b}}.
\newblock URL \url{https://arxiv.org/abs/2505.14362}.

\bibitem[Zhou et~al.(2025{\natexlab{a}})Zhou, An, Chi, Han, Rong, Zhang, Wang, Wang, Huang, Sheng, et~al.]{zhou2025roboreferspatial}
Enshen Zhou, Jingkun An, Cheng Chi, Yi~Han, Shanyu Rong, Chi Zhang, Pengwei Wang, Zhongyuan Wang, Tiejun Huang, Lu~Sheng, et~al.
\newblock Roborefer: Towards spatial referring with reasoning in vision-language models for robotics.
\newblock \emph{ArXiv preprint}, abs/2506.04308, 2025{\natexlab{a}}.
\newblock URL \url{https://arxiv.org/abs/2506.04308}.

\bibitem[Zhou et~al.(2025{\natexlab{b}})Zhou, Qiu, Chen, Wang, Yang, Xu, and Qiu]{zhou2025reinforced}
Guanghao Zhou, Panjia Qiu, Cen Chen, Jie Wang, Zheming Yang, Jian Xu, and Minghui Qiu.
\newblock Reinforced mllm: A survey on rl-based reasoning in multimodal large language models.
\newblock \emph{ArXiv preprint}, abs/2504.21277, 2025{\natexlab{b}}.
\newblock URL \url{https://arxiv.org/abs/2504.21277}.

\end{thebibliography}
\bibliographystyle{colm2026_conference}

\appendix
\section{More Results}
\label{sec:more results}

We demonstrate \model's training stability and efficiency in accuracy reward, length reward, format reward and structure reward.

\begin{figure}[h]
  \centering
\includegraphics[trim=0cm 0cm 0cm 0cm, clip, width=\linewidth]{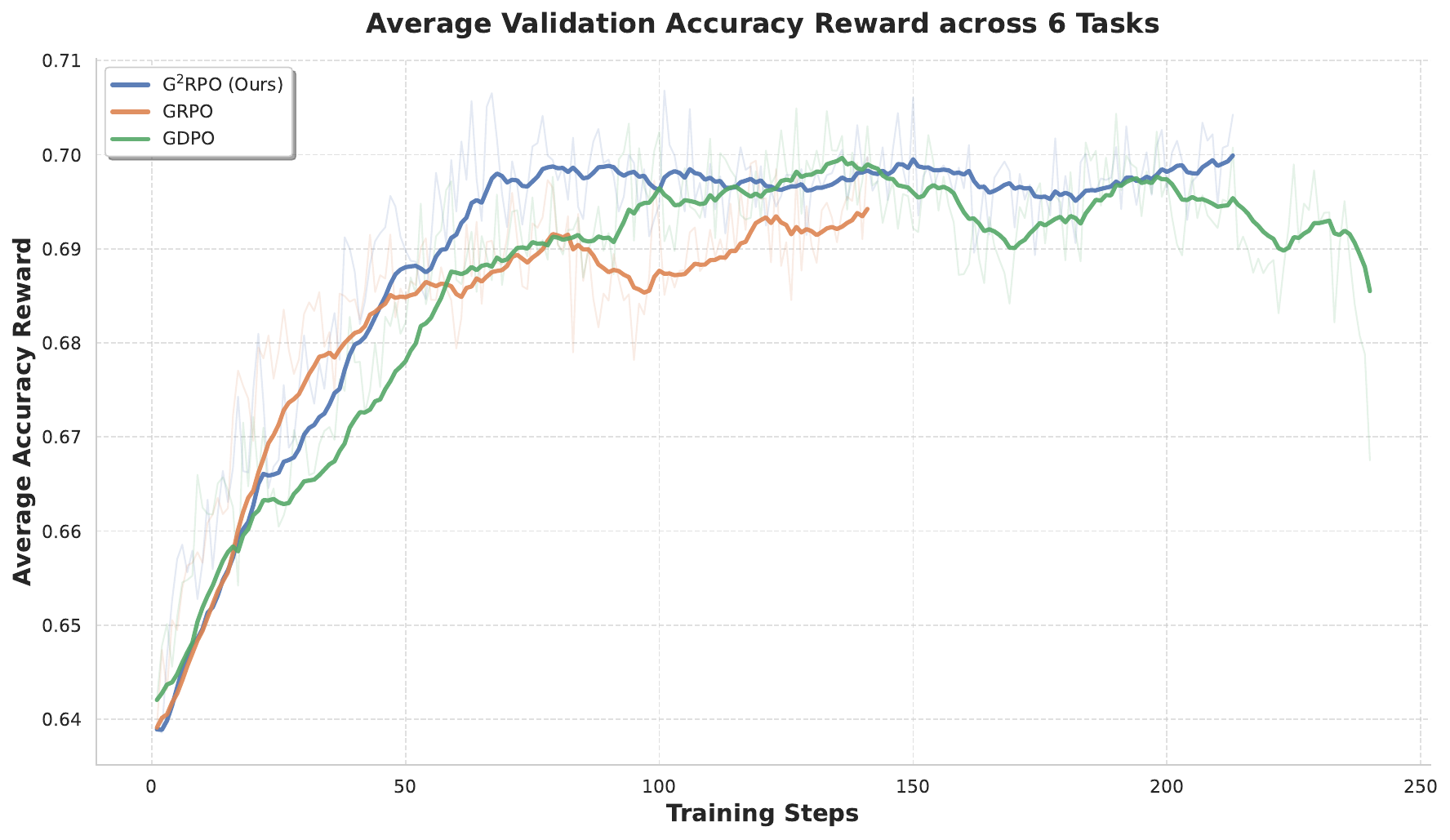}
  \caption{Average Accuracy Reward Comparison across all tasks on the Validation set during Training. \model demonstrates stable and superior performance overall.}
 \label{fig:training_compare}
 \vspace{-3mm}
\end{figure}

\textbf{Accuracy Reward.}
As showcased in Figure~\ref{fig:training_compare},
comparing to GRPO and GDPO baselines, \model demonstrates an early convergence at the first of the 100 training steps. While GRPO is oscillating between 0.685 and 0.695 and GDPO oscillate to lower accuracy reward around step 240, our method (\model) is consistently learning and improving its accuracy reward. 

\begin{figure}[h]
  \centering
\includegraphics[trim=0cm 0cm 0cm 0cm, clip, width=\linewidth]{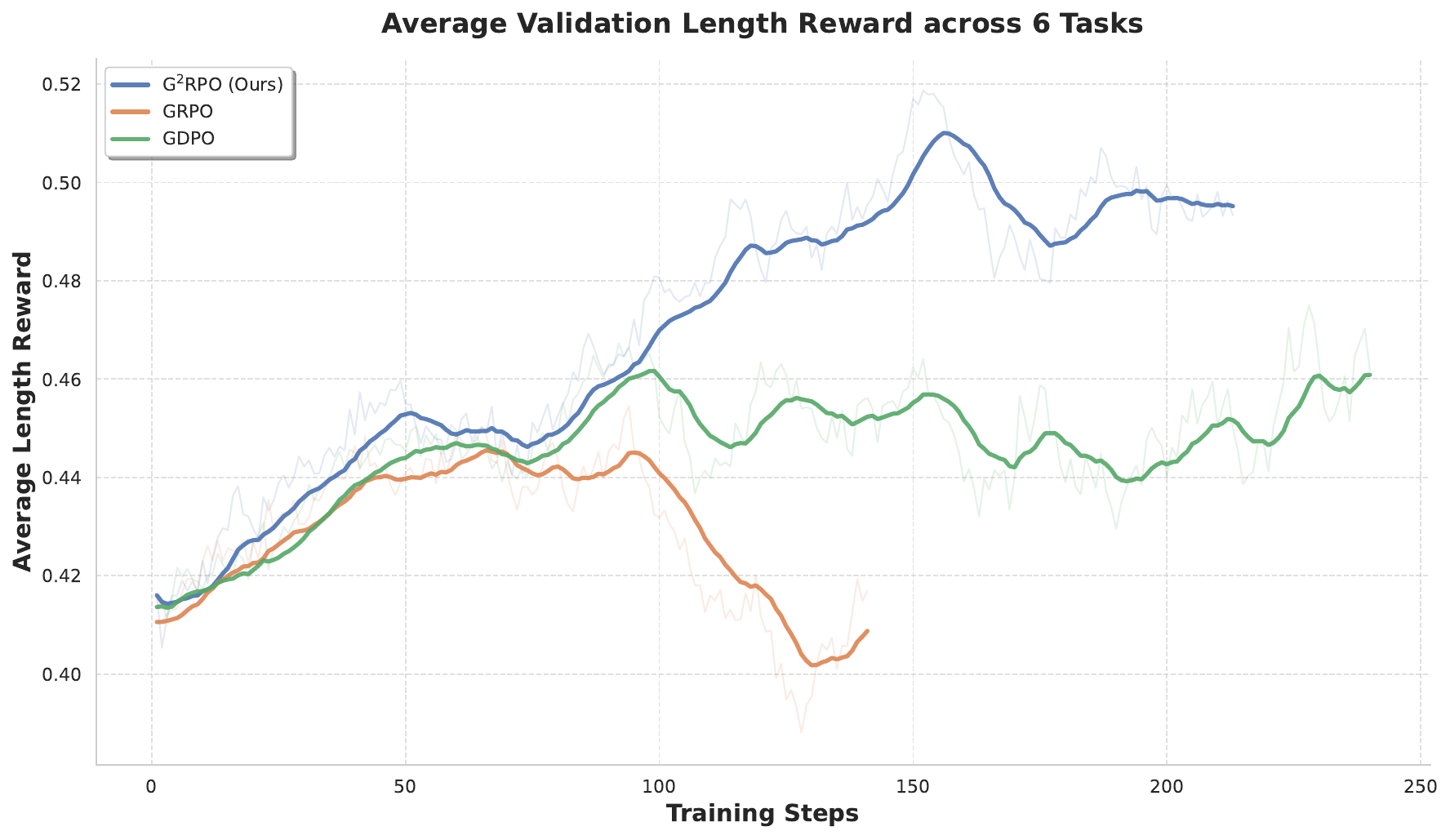}
  \caption{Average Length Reward Comparison across all tasks on the Validation set during Training. \model demonstrates stable and superior performance overall.}
 \label{fig:training_compare_length}
 \vspace{-3mm}
\end{figure}

\textbf{Length Reward.} As illustrated in Figure~\ref{fig:training_compare_length},
 \model consistently demonstrate significantly higher length reward than the two other baselines.

\begin{figure}[h]
  \centering
\includegraphics[trim=0cm 0cm 0cm 0cm, clip, width=\linewidth]{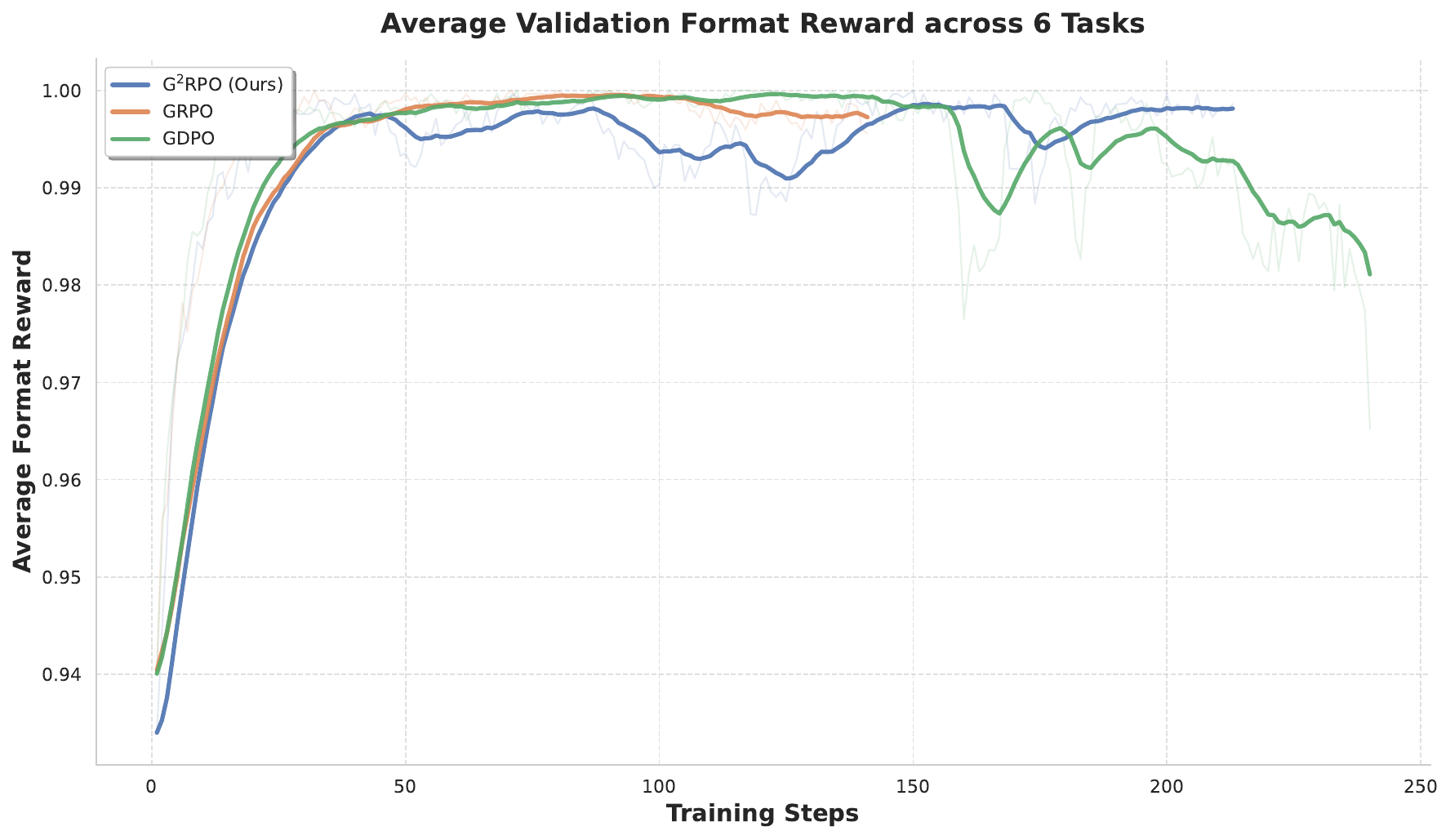}
  \caption{Average Format Reward Comparison across all tasks on the Validation set during Training. \model demonstrates stable and superior performance overall.}
 \label{fig:training_compare_format}
 \vspace{-3mm}
\end{figure}

\textbf{Format Reward.} 
Following~\cite{feng2025onethinkerallinonereasoningmodel}, we adopt format reward which enclose the thinking process with <thinking> and  </thinking> tokens and enclose the final answer with the <answer> and </answer> tokens. While GDPO has a  theoretical advantage maintaining a better format reward, it demonstrates highest format reward at the beginning but then converges to lower result, as illustrated in Figure~\ref{fig:training_compare_format}. \model maintains the best format reward at the end of the training.

\textbf{Structure Reward.} 
Following~\cite{feng2025onethinkerallinonereasoningmodel}, we also adopt structure reward, specifically for tasks that require a strict output structure. For example the grounding task output adopt the format of <answer>{"boxes": [a, b, c, d]}</answer>.
While GDPO has a theoretical advantage maintaining a better structure reward, similar to format reward result, it demonstrates highest structure reward at the beginning but then converges to lower result. \model maintains the best structure reward at the end of the training.

\begin{figure}[h]
  \centering
\includegraphics[trim=0cm 0cm 0cm 0cm, clip, width=\linewidth]{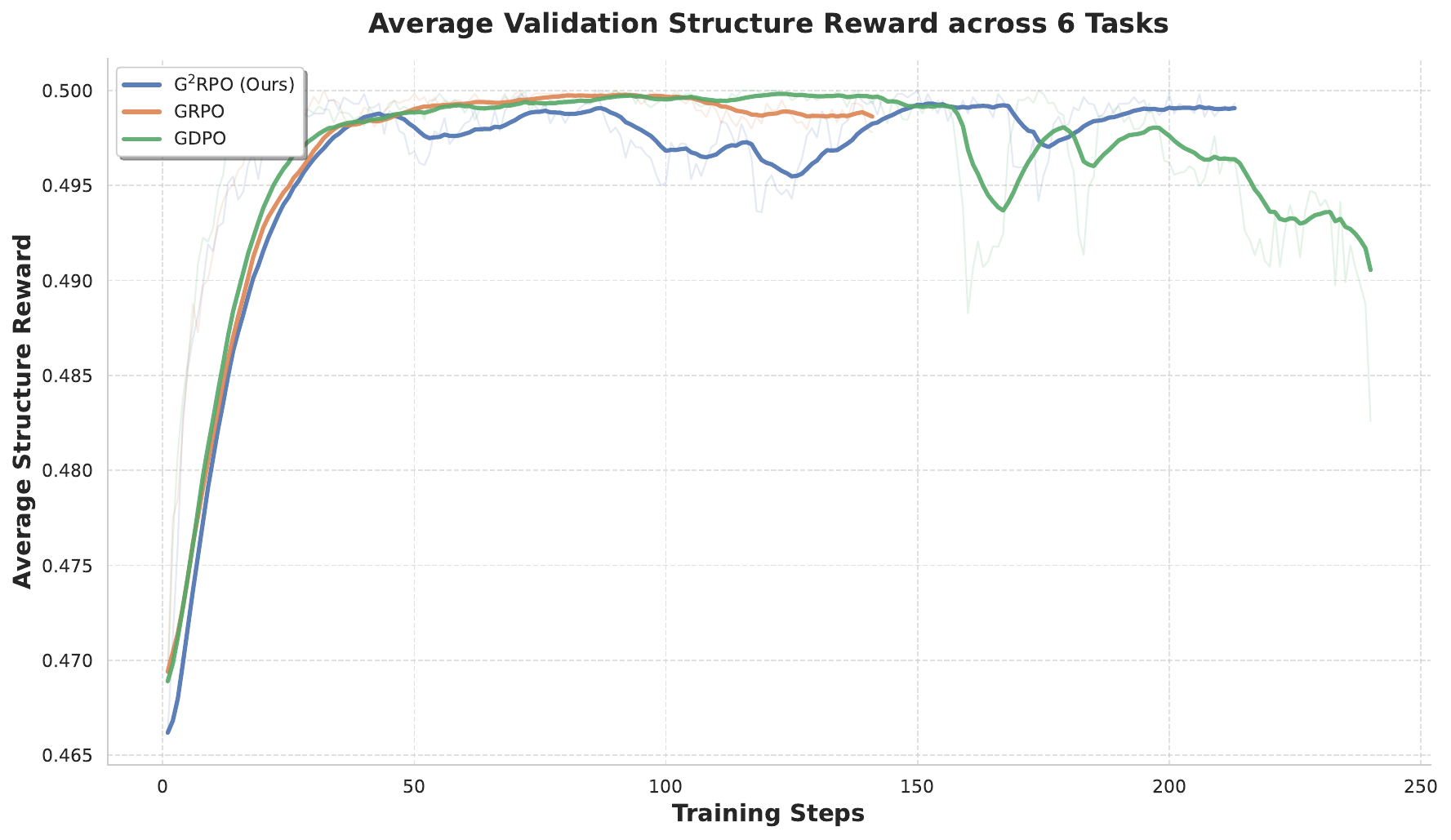}
  \caption{Average Structure Reward Comparison across all tasks on the Validation set during Training. \model demonstrates stable and superior performance overall.}
 \label{fig:training_compare_structure}
 \vspace{-3mm}
\end{figure}

\section{Policy Gradient Derivation of  \model}
\label{subsec:g_grpo_derivation}

To derive the policy gradient for our proposed \model, we begin with the fundamental expected return objective and systematically replace the linear standard advantage with our non-linear Optimal Transport advantage mapping.

\textbf{The Base Policy Gradient.}
For an autoregressive language model parameterised by $\theta$, given a query $q \sim \mathcal{D}$ and a response $y$, the standard policy gradient theorem dictates that the gradient of the expected return $\mathcal{J}(\theta)$ is driven by the advantage $A(q, y)$:
\begin{equation}
    \nabla_\theta \mathcal{J}(\theta) = \mathbb{E}_{q \sim \mathcal{D}, y \sim \pi_\theta} \left[ \sum_{t=1}^{|y|} \nabla_\theta \log \pi_\theta(y_t | q, y_{<t}) A(q, y) \right]
\end{equation}

\textbf{Group Sampling and Importance Sampling.} 
To stabilize training without a separate value network, we adopt the group-sampling strategy of GRPO. For a query $q$, we sample a group of $G$ responses $\mathcal{G} = \{y_1, \dots, y_G\}$ from the behavior policy $\pi_{\theta_\text{old}}$. To optimize the new policy $\pi_\theta$ using these trajectories, we introduce the token-level importance sampling ratio:
\begin{equation}
    r_{i,t}(\theta) = \frac{\pi_\theta(y_{i,t} | q, y_{i,<t})}{\pi_{\theta_\text{old}}(y_{i,t} | q, y_{i,<t})}
\end{equation}

\textbf{Injecting the Gaussian Advantage $\widehat{A}^{\model}$.} 
In standard GRPO, the advantage $\widehat{A}_i$ is computed via scalar standardization, which is vulnerable to heavy-tailed distributions and multi-task scale discrepancies. We replace this with our Gaussian Optimal Transport advantage, denoted as $\widehat{A}_i^{\model}$ 

For the sampled group's rewards $\mathcal{R}_{\tau} = \{R_1, \dots, R_G\}$, we calculate the rank-based empirical probability $p_i = \frac{\text{rank}(R_i) - 0.5}{G}$. The advantage is strictly mapped to the standard normal distribution $\mathcal{N}(0,1)$ using the inverse CDF $\Phi^{-1}$. Incorporating our tie-breaking strategy over the set of indices $\mathcal{K}_{R_i}$ sharing identical rewards, the advantage becomes:
\begin{equation}
    \widehat{A}_i^{\model} = \frac{1}{|\mathcal{K}_{R_i}|} \sum_{j \in \mathcal{K}_{R_i}} \sqrt{2} \operatorname{erfinv}(2p_j - 1)
\end{equation}
Because $\widehat{A}_i^{\model}$ perfectly matches the quantiles of $\mathcal{N}(0,1)$, it guarantees $\sum_i \widehat{A}_i^{\model} \approx 0$ and $\sum_i (\widehat{A}_i^{\model})^2 \approx 1$, inherently satisfying the zero-mean and unit-variance requirements for stable policy updates, while systematically mitigating the influence of reward outliers.

\textbf{The Clipped Surrogate Objective.} 
To prevent destructively large policy updates, we bind the policy update using a PPO-style clipping mechanism. By integrating our Gaussian advantage $\widehat{A}_i^{\model}$ into the surrogate objective, we define the formal loss function for \model:
\begin{equation}
\begin{aligned}
    \mathcal{J}_{\model}(\theta) &= \mathbb{E}_{q \sim \mathcal{D}, \{y_i\}_{i=1}^G \sim \pi_{\theta_\text{old}}} \Bigg[ \frac{1}{G} \sum_{i=1}^{G} \frac{1}{|y_i|} \sum_{t=1}^{|y_i|} \\
    &\quad \min \left( r_{i,t}(\theta) \widehat{A}_i^{\model}, \, \text{clip}\left( r_{i,t}(\theta), 1 - \varepsilon, 1 + \varepsilon \right) \widehat{A}_i^{\model} \right) \Bigg]
\end{aligned}
\end{equation}

\textbf{The Final Gradient Update.} 
During backpropagation, the gradient of our surrogate objective with respect to $\theta$ determines the parameter updates. Let $\mathcal{L}_{\text{clip}}^{i,t}(\theta)$ denote the inner clipped term. The empirical gradient is evaluated as:
\begin{equation}
    \nabla_\theta \mathcal{J}_{\model}(\theta) = \mathbb{E}_{q \sim \mathcal{D}, \{y_i\}_{i=1}^G \sim \pi_{\theta_\text{old}}} \left[ \frac{1}{G} \sum_{i=1}^{G} \frac{1}{|y_i|} \sum_{t=1}^{|y_i|} \nabla_\theta \mathcal{L}_{\text{clip}}^{i,t}(\theta) \right]
\end{equation}
where the gradient of the token-level loss evaluates to:
\begin{equation}
    \nabla_\theta \mathcal{L}_{\text{clip}}^{i,t}(\theta) =
    \begin{cases}
        \widehat{A}_i^{\model} \, r_{i,t}(\theta) \nabla_\theta \log \pi_\theta(y_{i,t} | q, y_{i,<t}) & \text{if } r_{i,t}(\theta) \text{ is not clipped} \\
        0 & \text{otherwise}
    \end{cases}
\end{equation}

\end{document}